\documentclass{article}
\usepackage[final,nonatbib]{neurips_2024}
\usepackage[utf8]{inputenc} 
\usepackage[T1]{fontenc}    
\usepackage{hyperref}       
\hypersetup{hidelinks=true}
\usepackage{url}            
\usepackage{booktabs}       
\usepackage{nicefrac}       
\usepackage{microtype}      
\usepackage{xcolor}         
\usepackage{listings}%
\usepackage{colortbl}
\usepackage{makecell}
\usepackage{subfigure}
\usepackage{graphicx}%
\usepackage{multirow}%
\usepackage{amsmath,amssymb,amsfonts}%
\usepackage{amsthm}%
\usepackage{mathrsfs}%
\usepackage{textcomp}%
\usepackage{algorithm}%
\usepackage{algpseudocode}%
\newcommand{\ie}{i.e.,\ }
\newcommand{\eg}{e.g.,\ }

\newcommand{\et}{\emph{et al.}\ }

\title{Towards Effective and Efficient Adversarial Defense with Diffusion Models for Robust Visual Tracking}

\author{
Long Xu$^1$, Peng Gao$^1$, Wen-Jia Tang$^1$, Fei Wang$^2$, Ru-Yue Yuan\\
$^1$Qufu Normal University\quad $^2$Harbin Institute of Technology Shenzhen\\
}

\begin{document}

\maketitle

\begin{abstract}
Although deep learning-based visual tracking methods have made significant progress, they exhibit vulnerabilities when facing carefully designed adversarial attacks, which can lead to a sharp decline in tracking performance. To address this issue, this paper proposes for the first time a novel adversarial defense method based on denoise diffusion probabilistic models, termed DiffDf, aimed at effectively improving the robustness of existing visual tracking methods against adversarial attacks. DiffDf establishes a multi-scale defense mechanism by combining pixel-level reconstruction loss, semantic consistency loss, and structural similarity loss, effectively suppressing adversarial perturbations through a gradual denoising process. Extensive experimental results on several mainstream datasets show that the DiffDf method demonstrates excellent generalization performance for trackers with different architectures, significantly improving various evaluation metrics while achieving real-time inference speeds of over 30 FPS, showcasing outstanding defense performance and efficiency. Codes are available at \url{https://github.com/pgao-lab/DiffDf}.\\
\textbf{Keywords:} Adversarial defense, adversarial attack, denosing, diffusion model, visual tracking.
\end{abstract}

\section{Introduction}\label{sec:1}

Visual tracking, as one of the core tasks in computer vision, plays a key role in practical application scenarios such as autonomous driving, intelligent surveillance, and human-computer interaction. In recent years, thanks to the rapid development of deep learning technology, the performance of visual tracking methods has significantly improved. However, existing deep learning-based trackers generally exhibit high vulnerability to adversarial attacks. Attackers can dramatically reduce the tracking accuracy and robustness by constructing carefully designed perturbations (\ie adversarial examples) that are difficult for the human vision system to detect, posing a serious threat to the security of practical application systems.

Adversarial attacks on visual tracking can be categorized into black-box and white-box attacks. White-box attacks fully utilize the internal information of the tracker, such as gradients, to iteratively optimize and generate highly targeted perturbations, thus disrupting tracking performance. However, white-box attacks are complex and challenging to meet real-time application requirements. In contrast, black-box attacks do not require access to the internal information of the tracker. They involve pre-training perturbations and directly applying them to the tracker, offering advantages in computational efficiency and real-time performance. However, black-box attacks tend to have weaker targeting and cannot maximize the attack effect. To address these two types of adversarial attacks, several defense methods have been proposed, such as adversarial training\cite{r23,r24}, input transformation\cite{r25,r26}, and feature enhancement\cite{r29,r30}. However, these methods struggle to achieve an effective trade-off between robustness, efficiency, and generalization ability. Adversarial training typically relies on the generation of a large number of examples and intensive gradient optimization, which results in huge training overhead and is prone to overfitting to specific attack types, making it less effective when handling unknown perturbations or transferring to new attack types. Input transformation methods, such as JPEG compression\cite{r25} or Gaussian smoothing\cite{r26}, have low computational costs, but they often damage the structural details and semantic consistency of the image, making them weak in tracking target perception under strong perturbations. Feature enhancement methods usually remove interference by inserting auxiliary modules into the intermediate feature space. While they show some recovery effect, they also introduce significant computational and parameter redundancy, making it difficult for them to meet real-time requirements. Particularly in visual tracking tasks, which are dynamic and continuous over multiple frames, existing methods face bottlenecks such as weak generalization, poor timeliness, and incomplete structural recovery, urgently requiring a defense mechanism that is robust and adaptable to the task at hand.

In recent years, denoising diffusion probabilistic models (DDPM)\cite{r1}, as an emerging generative framework, have demonstrated strong capabilities in image denoising and generation. Diffusion models generate data by gradually adding and removing noise, naturally possessing efficient denoising characteristics, offering new insights for defensing adversarial attacks. Previous studies such as GDMP\cite{r33} and Diffpure\cite{r34} have preliminarily confirmed the adversarial defense potential of diffusion models in image classification tasks, but their application in the more complex task of visual tracking has yet to be explored.

\begin{figure}[h!]
	\begin{center}
		\includegraphics[width=0.8\linewidth]{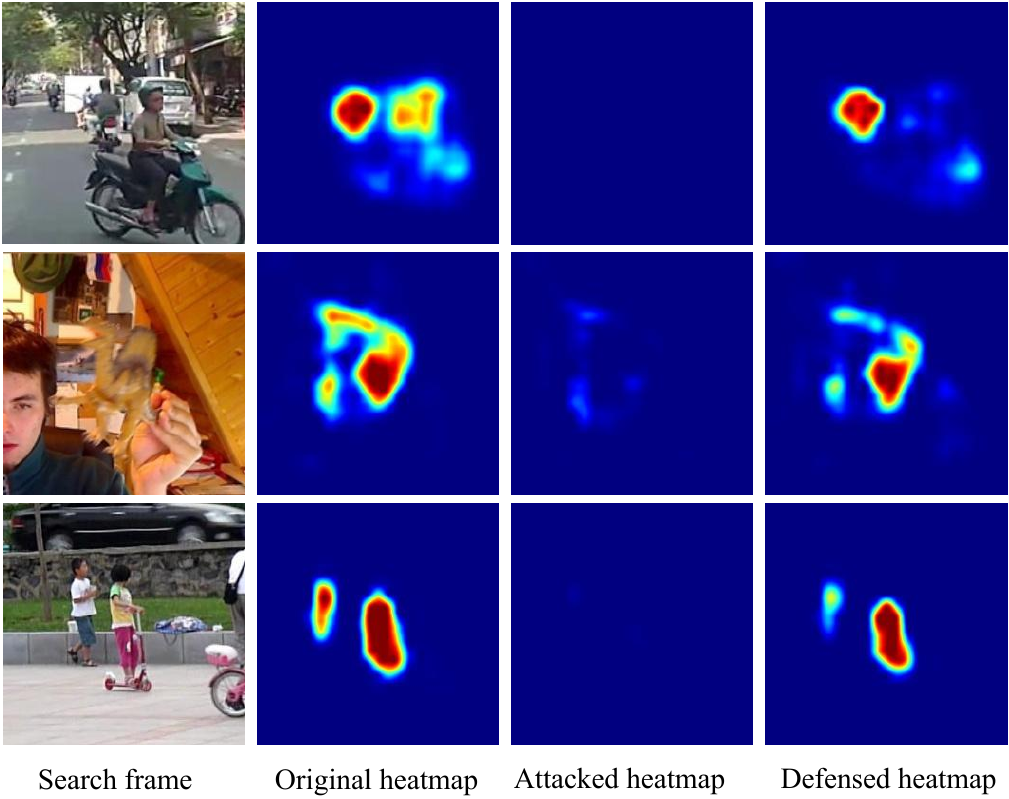}
	\end{center}
	\caption{Response map comparisons of the SiamRPN++\cite{r2} tracker before adversarial attack (using the black-box IoU attack method\cite{r19}), after adversarial attack, and after applying our proposed DiffDf method for defense.}
	\label{fig:1}
\end{figure}

To address the above issues, this paper proposes, for the first time, a multi-scale adversarial defense method based on DDPM, named DiffDf, aimed at enhancing the robustness of existing visual trackers against adversarial attacks. Specifically, we leverage the DDPM’s gradual denoising mechanism and introduce pixel-level reconstruction loss, semantic consistency constraints, and structural similarity loss to form multi-scale constraints, optimizing the quality of the restored images comprehensively. These three losses jointly impose constraints from image detail restoration, high-level semantic preservation, and local structural consistency: The pixel-level reconstruction loss is used to maintain the texture details of the image, the semantic consistency loss enhances the semantic robustness by aligning pre-trained features, and the structural similarity loss improves the ability to recover from complex structural deformations. This joint constraint builds a multi-scale loss mechanism for pixel, structure, and semantic co-learning during the diffusion process, providing a more stable and generalizable optimization objective for the defense method. Fig.\ref{fig:1} shows the response map comparisons of the SiamRPN++\cite{r2} tracker before adversarial attack (using the black-box IoU attack method\cite{r19}), after adversarial attack, and after applying our proposed DiffDf method for defense. The DiffDf method effectively restores the focus on the target object during tracking and even reduces attention to the distractors. This validates the effectiveness of DiffDf in feature recovery and anti-interference. Meanwhile, through extensive experimental analysis and ablation studies, we further explore the impact of key factors such as time step embedding mechanisms and the number of diffusion steps on model performance, systematically verifying the effectiveness and efficiency of the proposed DiffDf method. The main contributions of our study are as follows:
\begin{itemize}
	\item An adversarial defense method for visual tracking based on diffusion models is proposed, providing a general framework for removing adversarial perturbations.
	\item A multi-scale constraint mechanism is introduced, effectively balancing the pixel-level, semantic-level, and structural-level quality of the denoised images, significantly improving the defense performance.
	\item The effectiveness and efficiency of the proposed defense method are demonstrated through various experiments on multiple mainstream datasets and representative existing trackers.
\end{itemize}

The remainder of this paper is organized as follows: Section \ref{sec:2} provides a review of recent works relevant to the study. In Section \ref{sec:3}, we describe the proposed DiffDf method in detail. Section \ref{sec:4} presents the experimental results. Finally, Section \ref{sec:5} concludes the paper.

\section{Related works}\label{sec:2}

\subsection{Visual tracking}

Existing visual tracking methods can be divided into three categories: Siamese network-, discriminative learning-, and Transformer-based trackers. Siamese network-based trackers, pioneered by SiamFC\cite{r5}, achieve efficient tracking through weight-shared feature extraction and cross-correlation matching networks. Subsequent improvements include SiamRPN\cite{r6}, which incorporates the region proposal network (RPN) to optimize anchor regression; SiamRPN++\cite{r2}, which uses the deep ResNet architecture to enhance feature representation; and SiamMask\cite{r7}, which strengthens the recognition ability of deformable target objects through an additional segmentation network. Discriminative learning-based trackers improve robustness by online learning the appearance features of the target object, such as DiMP\cite{r3}, which introduces a discriminative learning framework to enhance the discrimination between the target object and background distractors; ATOM\cite{r8}, which optimizes target localization by maximizing overlap; and PrDiMP\cite{r9}, which further improves tracking accuracy by combining probabilistic regression. Transformer-based trackers break through the limitations of the local receptive field of traditional convolutional neural networks (CNNs). For instance, TransT\cite{r4} uses cross-attention to establish the feature association between the target template and search region; STARK\cite{r10} captures the motion pattern through spatiotemporal token encoding; and MixFormer\cite{r11} combines CNN and Transformer for multi-scale feature fusion. In addition, the field of pedestrian re-identification under occlusion has proposed various effective robust modeling strategies, offering new perspectives for designing adversarial defense strategies in visual tracking. Yan \et \cite{r46} leverages CLIP-guided fine-grained cross-modal feature alignment to mine more discriminative local features. Dong \et \cite{r47} establishes an adversarial training mechanism to enhance occlusion robustness through feature erasure, transformation, and noise injection. Dong \et \cite{r48} utilizes multi-view information integration and knowledge transfer strategies to restore the discriminative ability for occluded images. These methods provide important references for temporal tracking approaches in addressing challenges such as local occlusion and perturbation accumulation, particularly regarding feature stability and occlusion awareness.

Despite existing visual tracking methods have demonstrated excellent performance, vulnerabilities in different types of trackers have gradually become apparent when facing adversarial attacks. Siamese network-based trackers rely on the similarity-matching mechanism via cross-correlation operations and are very sensitive to feature space perturbations. For example, the anchor regression in SiamRPN\cite{r6} and SiamRPN++\cite{r2} are easily affected by slight shifts in the feature map, leading to regression errors. Due to its pixel-level prediction characteristics, the segmentation network of SiamMask\cite{r7} is more susceptible to interference from local adversarial perturbations. While discriminative learning-based trackers improve robustness by online learning appearance features, their online updating mechanism can be misled by adversarial examples. For instance, the foreground-background binary classifiers in DiMP\cite{r3} and ATOM\cite{r8} may generate incorrect confidence distributions under adversarial perturbations. Long-range dependencies learned by Transformer-based trackers may amplify the impact of adversarial perturbations. For example, the cross-attention weights in TransT\cite{r4} can be misled to non-target regions, and the spatiotemporal encoding module in STARK\cite{r10} might generate erroneous historical state transfers under sequential disturbances. Furthermore, offline-trained feature extraction networks, such as ResNet and Swin Transformer, exhibit gradient saturation regions, where adversarial attackers can induce sequential distortions in the feature space with minor input perturbations. Meanwhile, the lagging target template updating mechanism in existing tracking methods can cause adversarial attacks to accumulate over multiple frames. In addition, visual tracking methods possess frame-by-frame temporal continuity, and their feature extraction mechanisms are often susceptible to perturbations. Once the feature representation is disrupted, even a minimal level of perturbation can lead to matching failures and tracking drift, which accumulates and amplifies over subsequent frames, thereby significantly magnifying the impact of the attack. Therefore, the characteristics of input sensitivity and temporal propagation in visual tracking tasks make them a high-risk domain for adversarial attacks, urgently requiring effective defense mechanisms that ensure both robustness and temporal stability.

Although multi-stage fusion strategies\cite{r11} or dynamic target template updating\cite{r10} have been introduced into the visual tracking community to improve performance, they have not fundamentally solved the vulnerability of existing visual tracking methods to adversarial attacks. This challenge is especially prominent in complex scenarios where attackers can induce persistent errors in critical phases such as target localization and scale estimation by constructing adversarial perturbations that are hard for the human vision system to recognize, posing a significant threat to practical application scenarios.

Moreover, although existing visual tracking methods perform excellently in most scenarios, they generally rely on feature similarity matching and historical frame modeling, which makes them prone to semantic shifts and error accumulation when faced with perturbations. This indicates that trackers of different architectures share a common vulnerability in terms of robustness, highlighting the urgent need to develop defense mechanisms with perturbation awareness and temporal recovery capabilities.

\subsection{Adversarial attacks}

Adversarial attacks, as an important research topic in deep learning model security, were first proposed by Szegedy \et in 2014\cite{r13} and have made groundbreaking progress in image classification. By adding carefully designed small perturbations to the input image, the classification accuracy of the model can be significantly reduced, or even cause the model to output completely incorrect predictions. This reveals the vulnerability of deep learning models when facing adversarial attacks.

Typical adversarial attack methods include white-box attacks based on gradient information (such as FGSM\cite{r14}, PGD\cite{r15}, and C\&W\cite{r16}) and black-box attacks that do not rely on internal model information (such as ZOO\cite{r17} and Boundary Attack\cite{r18}).
In recent years, adversarial attacks aimed at visual tracking have gradually become essential to reveal model vulnerabilities. Adversarial attacks have expanded from traditional static image classification tasks to dynamic visual tracking, where researchers generate imperceptible adversarial perturbations to disrupt tracking accuracy and robustness. In white-box attacks, the attacker must have complete access to the model's structural parameters and gradient information to directly optimize adversarial perturbations based on the loss function. In contrast, black-box attacks cannot access the internal information of the model, and the attacker must rely on query mechanisms or transfer models to carry out the attack. Moreover, to reduce dependence on current frame information, some black-box attack methods propose a strategy of pre-training perturbations offline and then injecting noise online, allowing the attack to be executed without real-time computation, significantly reducing computational overhead. Furthermore, attackers must balance the perturbation magnitude, query count, and success rate, striking a trade-off between attack efficiency and perceptibility to develop practical and covert adversarial strategies.

Current adversarial attack methods for visual tracking can be divided into two categories based on how the attacks are executed: (1) online attacks, which generate targeted perturbations based on dynamic information from the current frame during the tracking process, but require real-time computation and are less efficient\cite{r19,r20,r21}; (2) offline attacks, which generate generic adversarial perturbations through pre-training and directly apply them during the tracking process\cite{r22}, making them more computationally efficient, though the generated perturbations are not as targeted as those from online attacks.

\subsection{Adversarial defenses}

Existing adversarial defense methods are mainly divided into three categories: adversarial training, input preprocessing, and feature enhancement. Adversarial training introduces adversarial examples during model training, enabling the model to actively recognize and resist adversarial perturbations. TRADES\cite{r23} adversarial training method significantly improves model robustness by optimizing the loss function for the worst-case scenario. MART\cite{r24} further refines the optimization strategy, effectively enhancing the stability of adversarial training. Input preprocessing methods reduce or eliminate the impact of adversarial perturbations by applying transformations or filters to the input data. JPEG compression\cite{r25} and Gaussian smoothing\cite{r26} mitigate adversarial perturbations by reducing detail information in the image. At the same time, random cropping\cite{r27} and image quantization\cite{r28} have also been proven to effectively alleviate the impact of adversarial perturbations. Feature enhancement methods strengthen the representation capability of target models to improve their resistance to adversarial perturbations. For instance, feature normalization\cite{r29} and attention mechanisms\cite{r30} enhance the significance of target features, boosting model robustness, while multi-scale feature fusion\cite{r31} and adversarial feature distillation\cite{r32} further broaden the application of feature enhancement methods. However, the aforementioned traditional methods often have limitations such as high computational cost, limited defense effectiveness, and insufficient generalization ability, making them difficult to fully adapt to complex scenarios. In addition to general adversarial defense methods for computer vision tasks, recent works have also explored defense strategies tailored to specific tasks and domain adaptation. Zeng \et\cite{r41} proposed a federated learning-based adversarial attack and defense framework customized for industrial control systems, which enhances the robustness of distributed devices. In SAR image classification, Wei \et\cite{r42} introduced MoAR CNN, a multi-objective adversarial CNN that balances classification accuracy and robustness. Zhang \et\cite{r43} proposed DoFA, a dual-objective feature attribution method aimed at detecting adversarial examples by analyzing feature contributions. Additionally, Zeng \et\cite{r44} introduced DACO-BD, a compositional data augmentation method for defending against backdoor attacks in SAR classification. These works demonstrate that task-specific constraints and domain knowledge can significantly enhance the robustness of adversarial defenses. In summary, current works gradually focus on domain adaptation and task-specific adversarial defense designs, demonstrating a trend from general frameworks to application-oriented solutions. This development provides new research perspectives for enhancing adversarial robustness in high-risk tasks like visual tracking. Besides, recent studies in RGB-T saliency detection have also provided valuable insights into enhancing the robustness of feature fusion in complex scenarios. For example, the ConTriNet framework \cite{r45} proposed by Tang \et constructs three information streams to model modality-specific and complementary features separately and effectively reduces the impact of inter-modality discrepancies and input defects through a dynamic aggregation mechanism. The multi-branch fusion structure and dynamic interaction strategy proposed in this method have provided beneficial inspiration for designing our multi-scale feature defense architecture.

Most existing adversarial defense methods originate from static classification tasks and lack specialized modeling capabilities for tracking-specific characteristics, such as temporal continuity and multi-frame disturbance propagation. Furthermore, current defense methods often involve trade-offs in computational complexity, generalization ability, or recovery fidelity, making it difficult to achieve an ideal defense strategy that balances both practicality and effectiveness. In recent years, DDPM\cite{r1}, as an emerging generative model, has shown distinct advantages in adversarial defense for image classification tasks, providing new ideas to address the limitations of traditional adversarial defense methods. Diffusion models generate high-quality images efficiently and stably by gradually adding and removing noise, making them naturally suitable for removing adversarial perturbations. GDMP\cite{r33} achieves semantically consistent adversarial defense by guiding the diffusion model’s generation process, effectively eliminating adversarial perturbations via a step-by-step reverse denoising process while maintaining the semantic integrity of the image. Diffpure\cite{r34} utilizes a multi-stage denoising strategy to effectively separate adversarial perturbations from semantic features, thereby efficiently restoring the original image semantics. Additionally, DMID\cite{r35} combines adaptive embedding and dynamic weight allocation mechanisms, further enhancing the defense ability of the target model against challenging adversarial examples by performing multi-scale feature extraction.

Inspired by the studies mentioned above, this paper applies the progressive denoising mechanism of diffusion models to adversarial defense in visual tracking for the first time and proposes a multi-scale adversarial defense method based on DDPM. This method significantly improves the robustness of existing visual tracking methods by comprehensively considering pixel reconstruction, semantic preservation, and structural consistency.

\section{Methodology}\label{sec:3}

\subsection{Denoising diffusion probabilistic model}

DDPM transforms the data distribution into a Gaussian distribution and achieves data reconstruction by defining a forward diffusion process and a reverse denoising Markov chain process.

\subsubsection{Forward diffusion}

The forward diffusion process gradually transforms the original data distribution $q(x_0)$ into an isotropic Gaussian distribution through time steps $t=\{1,2,\dots,T\}$. The noise-adding process at each time step $t$ is defined as:
\begin{equation}\label{eq:1}
	q(x_t|x_{t-1})=\mathcal{N}\left(x_t;\sqrt{1-\beta_t}x_{t-1},\beta_t\mathbf{I}\right),
\end{equation}
where $\beta_t$ is the noise scheduling parameter, usually using linear or cosine scheduling strategies, and $\mathbf{I}$ is the identity matrix with the same dimension as the input data $x_{t-1}$. By recursively expanding this, the noise data at any time step $t$ can be directly calculated from the original data $x_0$. Introducing $\alpha_t=1-\beta_t$ and the cumulative product $\bar{\alpha}_t=\prod_{s=1}^t\alpha_s$, a closed-form solution is obtained:
\begin{equation}\label{eq:2}
	x_t=\sqrt{\bar{\alpha}_t}x_0+\sqrt{1-\bar{\alpha}_t}\epsilon,
\end{equation}
where $\epsilon\sim\mathcal{N}(0,\mathbf{I})$ represents the standard Gaussian noise injected at each time step. When $t=1$, $x_1=\sqrt{\alpha_1}x_0+\sqrt{1-\alpha_1}\epsilon_1$ can be computed, and by recursively substituting into $t=2$, the noise terms are combined. Using the linearity property of Gaussian distributions, we obtain:
\begin{equation}\label{eq:3}
	x_2=\sqrt{\alpha_2(1-\alpha_1)}\epsilon_1+\sqrt{1-\alpha_2}\epsilon_2,
\end{equation}
where $x_2\sim\mathcal{N}(0,[1-\alpha_1\alpha_2]\mathbf{I})$. This process extends to $t$ steps, yielding the Eq.\ref{eq:2}. When $T\to\infty$, $\alpha_T \to 0$, and at this point, $x_T\sim\mathcal{N}(0,\mathbf{I})$ indicating that the data has completely degenerated into Gaussian noise.

\subsubsection{Reverse denoising}

The reverse process aims to gradually recover the original data $x_0$ from the noise distribution $x_T\sim\mathcal{N}(0,\mathbf{I})$. This process is defined as a Markov chain:
\begin{equation}\label{eq:4}
	p_\theta(x_{0:T})=p(x_T)\prod_{t=1}^Tp_\theta\left(x_{t-1}|x_t\right),
\end{equation}

The conditional distribution of each time step is given by:
\begin{equation}\label{eq:4-1}
	p_\theta(x_{t-1}|x_t)=\mathcal{N}\left(x_{t-1};\mu_\theta(x_t,t),\Sigma_\theta(x_t,t)\right),
\end{equation}
where $\mu_\theta(x_t,t)$ and $\Sigma_\theta(x_t,t)$ represent the mean and variance learned by the model. By applying Bayes theorem, the conditional mean of the actual reverse process is:
\begin{equation}\label{eq:5}
	\tilde{\mu}_t=\frac{\sqrt{\overline{\alpha}_t-1}\beta_t}{1-\overline{\alpha}_t}x_0+\frac{\sqrt{\alpha_t}(1-\overline{\alpha}_{t-1})}{1-\overline{\alpha}_t}x_t,
\end{equation}
where $\tilde{\mu}_t$ is the mean of the actual reverse process derived by Bayes theorem, and $x_0=\left(x_t-\sqrt{1-\bar{\alpha}_t}\epsilon\right)/\sqrt{\bar{\alpha}_t}$. After substituting into the Eq.\ref{eq:5}, the equation simplifies to:
\begin{equation}\label{eq:6}
	\mu_0(x_t,t)=\frac{1}{\sqrt{\alpha_t}}\left(x_t-\frac{\beta_t}{\sqrt{1-\bar{\alpha}_t}}\epsilon_\theta(x_t,t)\right),
\end{equation}
where $\epsilon_\theta(\cdot)$ is a neural network model to predict noise level, and $\epsilon_\theta(x_t,t)$ indicates the predicted noise component used to approximate the noise $\epsilon$ injected during the forward process. The variance is typically fixed at $\Sigma_{\theta}=\beta_{t}\mathbf{I}$ or $\Sigma_{\theta}=\frac{1-\bar{\alpha}_{t-1}}{1-\bar{\alpha}_{t}}\beta_{t}\mathbf{I}$, avoiding the learning of a complex covariance matrix to simplify the training process.

\subsubsection{Optimization objective}

The training objective of DDPM is to maximize the variational lower bound (VLB) of the data log-likelihood, which is expressed as:
\begin{equation}\label{eq:7}
	\mathcal{L}_\text{VLB}=\mathbb{E}_q\left[\log p_\theta\left(x_0|x_1\right)-\sum_{t=2}^TD_\text{KL}\left(q(x_{t-1}|x_t,x_0)\|p_\theta(x_{t-1}|x_t)\right)\right],
\end{equation}

Substituting Eqs.\ref{eq:5} and \ref{eq:6} leads to the simplification of the KL divergence term:
\begin{equation}\label{eq:8}
	\mathcal{L}_t=\mathbb{E}_{x_0,\epsilon}\left[\frac{\beta_t^2}{2\sigma_t^2\alpha_t(1-\bar{\alpha}_t)}\|\epsilon-\epsilon_\theta(x_t,t)\|^2\right],
\end{equation}
where $\sigma_t^2$ is defined as the variance parameter associated with the relevant time step, typically taken as $\sigma_t^2=\beta_t$ or $\sigma_t^2=\frac{1-\bar{\alpha}_{t-1}}{1-\bar{\alpha}_t}\beta_t$. To simplify training, Ho \et\cite{r1} proposed ignoring the weight coefficients and directly optimizing the noise prediction error:
\begin{equation}\label{eq:9}
	\mathcal{L}_\text{simple}=\mathbb{E}_{t,x_0,\epsilon}[\|\epsilon-\epsilon_\theta(x_t,t)\|^2],
\end{equation}

This objective function avoids complex weight calculations, and experiments show that its training stability is significantly better than the original VLB objective. During training, the time step $t$ is randomly sampled from a uniform distribution $\mathtt{Uniform}(1, T)$, and the noise $\epsilon$ is generated from a standard Gaussian distribution. The noisy data $x_t$ is computed via Eq.\ref{eq:2}.

\subsection{Adversarial defense with constrained optimization}

Adversarial attacks are typically defined as an optimization problem where the goal is to add adversarial perturbation $\delta$ to a clean image $x$ to maximize the loss function $\mathcal{L}(x, y)$ of the target model $\mathcal{M}(\cdot)$, such that the adversarial example $x^\prime$ remains within a predefined perturbation constraint space $\mathcal{B}(x)$:
\begin{equation}\label{eq:10}
	x^\prime = \arg\max_{x^\prime\in \mathcal{B}(x)} \mathcal{L}(\mathcal{M}(x^\prime), y),
\end{equation}
where $x^\prime=x+\delta$ indicates the adversarial example and $y$ is the ground-truth label of $x$. The constraint space $\mathcal{B}(x)$ is typically defined as $\|\delta\|_p \leq \epsilon$, where $p \in \{1, 2, \infty\}$ indicates the type of norm, and $\epsilon$ is the upper bound on the perturbation. Based on this, adversarial defense can be defined in two different approaches.

The first approach is to directly remove the adversarial perturbation by having the defense model $\mathcal{D}(\cdot)$ predict and remove the adversarial perturbation $\delta$, minimizing the loss function between the denoised example $\tilde{x}$ and the clean image $x$,
\begin{equation}\label{eq:11}
	\tilde{x}=\arg\min_{\tilde{x}\in \mathcal{D}(x^{\prime})}\mathcal{L}(\tilde{x},x),
\end{equation}

In this approach, the goal of the defense model $\mathcal{D}(\cdot)$ is to find a mapping such that the denoised example $\tilde{x}$ is as close as possible to the clean image $x$, thereby eliminating the impact of the adversarial perturbation on the performance.

The second approach is to optimize the task loss of the denoised example $\tilde{x}$. That is, the defense model $\mathcal{D}(\cdot)$ generates a denoised example $\tilde{x}$ that minimizes the loss between the predictions of the target model $\mathcal{M}(\cdot)$ on $\tilde{x}$ and the ground-truth label $y$:
\begin{equation}\label{eq:12}
	\tilde{x}=\arg\min_{\tilde{x}\in \mathcal{D}(x^{\prime})}\mathcal{L}(\mathcal{M}(x^{\prime}),y),
\end{equation}

In this approach, the defense model not only focuses on whether the denoised example is close to the clean image but also further optimizes its performance on downstream tasks such as classification, detection, and tracking, ensuring that the denoised example can still be correctly inferred by the target model $\mathcal{M}(\cdot)$.

\subsection{Adversarial defense with diffusion model}

We propose a defense method based on DDPM, modeling the adversarial defense task as a step-by-step process of perturbation removal. This approach ensures that the denoised examples are close to the clean images at the pixel level and effectively recover their semantic information for visual tracking. By integrating a progressive diffusion denoising mechanism with a perceptual loss optimization strategy, the method can adapt to various adversarial attacks and maintain high denoising effectiveness and stability across different perturbation levels. The core of our method involves transforming the denoising process of adversarial examples into inverse learning of noise prediction and gradual denoising. By introducing a multi-scale constraint strategy during the diffusion process, it jointly optimizes pixel-level reconstruction, semantic feature preservation, and local structure consistency, ultimately achieving efficient removal of adversarial perturbations and precise restoration of target-specific features. The proposed DiffDf is designed as an input-level preprocessing defense model. By modeling the image restoration process as a gradual denoising procedure, DiffDf effectively improves the image quality of adversarial examples without altering the structure of the original tracker. DiffDf demonstrates strong versatility and generalization, significantly enhancing the tracking robustness under adversarial perturbations.

Specifically, for a given adversarial example $x^\prime=x+\delta$ (with $x$ as the clean image and $\delta$ as the adversarial perturbation), we degrade it into a pure noise distribution by progressively adding Gaussian noise. The forward diffusion process is defined as:
\begin{equation}\label{eq:13}
	q(x_t|x_{t-1})=\mathcal{N}(x_t;\sqrt{\alpha_t}x_{t-1},(1-\alpha_t)\mathbf{I}),\quad t=1,2,\ldots,T,
\end{equation}
where $\alpha_t = 1-\beta_t$, $\beta_t$ is the noise scheduling parameter, and $\mathbf{I}$ is the identity matrix. By recursive expansion, noise data at any time step $t$ can be directly calculated from the initial adversarial example $x^\prime$ as,
\begin{equation}\label{eq:14}
	x_t=\sqrt{\bar{\alpha}_t}x^{\prime}+\sqrt{1-\bar{\alpha}_t}\epsilon,\quad\epsilon\sim\mathcal{N}(0,\mathbf{I}),
\end{equation}
where $\bar{\alpha}_t=\prod_{s=1}^t\alpha_s$.

In the reverse process, the neural network $\epsilon_\theta$ predicts the injected noise and gradually restores the clean image:
\begin{equation}\label{eq:15}
	p_\theta(x_{t-1}|x_t)=\mathcal{N}(x_{t-1};\mu_\theta(x_t,t),\Sigma_t \mathbf{I}),
\end{equation}
where $\Sigma_t$ is the fixed variance parameter (typically set as $\Sigma_t = \beta_t \mathbf{I} $), and the mean function $\mu_\theta$ is predicted by a U-Net and can be expressed as:
\begin{equation}\label{eq:16}
	\mu_\theta(x_t,t)=\frac{1}{\sqrt{\alpha_t}}\left(x_t-\frac{\beta_t}{\sqrt{1-\bar{\alpha}_t}}\epsilon_\theta(x_t,t)\right),
\end{equation}

During the training phase, adversarial examples at each batch are randomly sampled for time steps, and corresponding intensity noise is added according to Eq.\ref{eq:14}. The goal of the network model $\epsilon_\theta$ is to predict the injected noise, \ie minimize Eq.\ref{eq:9}. By training with random time steps, the network model learns the ability to recover clean images from any perturbation level, enhancing robustness against adversarial perturbations of varying intensities.

To balance pixel-level reconstruction and semantic consistency and restore image features at different levels, we introduce multi-scale constraints on top of the standard noise prediction loss. To enhance robustness without compromising the model's performance on clean inputs, we have designed a multi-scale loss constraint strategy in DiffDf, which includes pixel-level reconstruction, semantic consistency, and structural similarity. This combination not only suppresses perturbations but also improves the output image's fidelity and task awareness in non-attacked scenarios, effectively maintaining the performance of both clean images and adversarial examples.

\paragraph{Pixel-level reconstruction loss}Constrains the $\mathcal{L}_2$ distance between denoised example $\tilde{x} = \mathcal{D}(x^\prime)$ and the clean image $x$ as:
\begin{equation}\label{eq:16}
	\mathcal{L}_\text{pixel}=\mathbb{E}_{x,\delta,t}[\|\epsilon_\theta(x_t,t)-\epsilon_t\|^2],
\end{equation}
where $\epsilon_t$ is the random noise injected during the forward diffusion process, and $x_t$ is calculated via reparameterization as $x_t=\sqrt{\overline{\alpha}_t}x^{\prime}+\sqrt{1-\overline{\alpha}_t}\epsilon_t$.

\paragraph{Semantic consistency loss}Minimizes the distance in the feature space using the $\mathtt{conv4\_x}$ layer feature extractor $\phi(\cdot)$ from a pre-trained ResNet50:
\begin{equation}\label{eq:17}
	\mathcal{L}_\text{semantic}=\mathbb{E}_{x,\delta}[\|\phi(\mathcal{D}_\theta(\tilde{x}))-\phi(x)\|^2],
\end{equation}
where $\phi(\cdot)$ is used to compute the loss, ensuring denoised examples maintain consistency in high-level visual tasks like category discrimination.

\paragraph{Structural similarity loss}Optimizes local structural similarity through the SSIM metric:
\begin{equation}\label{eq:18}
	\mathcal{L}_\text{ssim}=1-\frac{(2\mu_\mathcal{D}\mu_x+C_1)(2\sigma_{\mathcal{D}x}+C_2)}{\left(\mu_\mathcal{D}^2+\mu_x^2+C_1\right)\left(\sigma_\mathcal{D}^2+\sigma_x^2+C_2\right)},
\end{equation}
where $\mu_\mathcal{D}$ and $\mu_x$ denote the local means of the denoised example and the clean image, respectively, $\sigma_{\mathcal{D}x}$ is the covariance, and $C_1 = 0.01$, $C_2 = 0.03$ are stability constants calculated using an 11$\times$11 Gaussian sliding window with a standard deviation of 1.5.

The pixel-level reconstruction loss constrains the retention of low-level details, helping preserve the texture and contours of the image. The semantic consistency loss, based on the pre-trained ResNet-50, calculates the alignment of high-level semantic features to ensure that semantic information is not disrupted. The structural similarity loss focuses on local structural consistency, enhancing the robustness against geometric distortions. During training, the pixel-level loss optimizes the model by guiding it to minimize the error between the predicted and the injected perturbation. The semantic loss, extracted from the conv4\_x layer features of the pre-trained ResNet-50, calculates semantic discrepancies. The structural similarity loss operates on the local SSIM between the reconstructed and clean images. These three losses jointly form a multi-scale constraint mechanism, stabilizing the diffusion process and improving the generalization ability against multiple types of perturbations.

The final optimization objective is a weighted combination form:
\begin{equation}\label{eq:19}
	\mathcal{L}_\text{total}=\mathcal{L}_\text{simple}+\lambda_1\mathcal{L}_\text{pixel}+\lambda_2\mathcal{L}_\text{semantic}+\lambda_3\mathcal{L}_\text{ssim},
\end{equation}
where $\lambda_1$, $\lambda_2$, and $\lambda_3$ are loss weight hyperparameters.

This method achieves adversarial perturbation separation through a standard diffusion framework without relying on conditional inputs. At the same time, the multi-scale losses ensure consistency of the denoised image at pixel, feature, and structural levels.

\subsection{Network architecture}

We employ a symmetric encoder-decoder architecture based on U-Net to implement our proposed DiffDf method. The encoder consists of 4 down-sampling modules, each comprised of two 3$\times$3 convolutional layers with a stride of 1, a GroupNorm layer with 32 groups, and a SiLU activation function. Each down-sampling stage reduces the input resolution by half through a convolutional layer with a stride of 2. At the same time, the number of channels progressively doubles (\ie 64$\rightarrow$128$\rightarrow$256$\rightarrow$512) to gradually extract multi-scale features and capture the global patterns of adversarial perturbations. The decoder restores the resolution using transposed convolutional layers with a stride of 2 and employs skip connections to concatenate features channel-wise from corresponding stages of the encoder, forming a symmetric U-shaped network architecture. The skip connection mechanism effectively integrates low-level geometric details (such as edges and textures) with high-level semantic features, mitigating the issue of information loss in deep networks and ensuring that the denoised example maintains high fidelity while eliminating adversarial perturbations.

The key to the diffusion model is modeling the denoising process at different noise levels. To this end, the network incorporates a time step embedding mechanism, which converts discrete time step information $t\in\{1,2,\ldots,T\}$ into continuous feature vectors. Specifically, the time step $t$ is first mapped into a high-dimensional vector through sinusoidal positional encoding, then adjusted in dimension through two fully connected layers, and finally added element-wise to the feature maps at each stage of the encoder. This design allows the model to dynamically adjust convolutional kernel weights according to the current noise level, accommodating the gradual denoising process from high-level noise (when $t$ is close to 1000) to low-level noise (when $t$ is close to 1). For instance, in the early stages (high-level noise), the model focuses on suppressing adversarial perturbations; in the later stages (low-level noise), it emphasizes detail reconstruction. Input images are uniformly resized to a resolution of 256$\times$256 using bilinear interpolation, and pixel values are normalized to the range $[-1,1]$ to fit the standard input format of the diffusion model.

\begin{figure}[h!]
	\begin{center}
		\includegraphics[width=\linewidth]{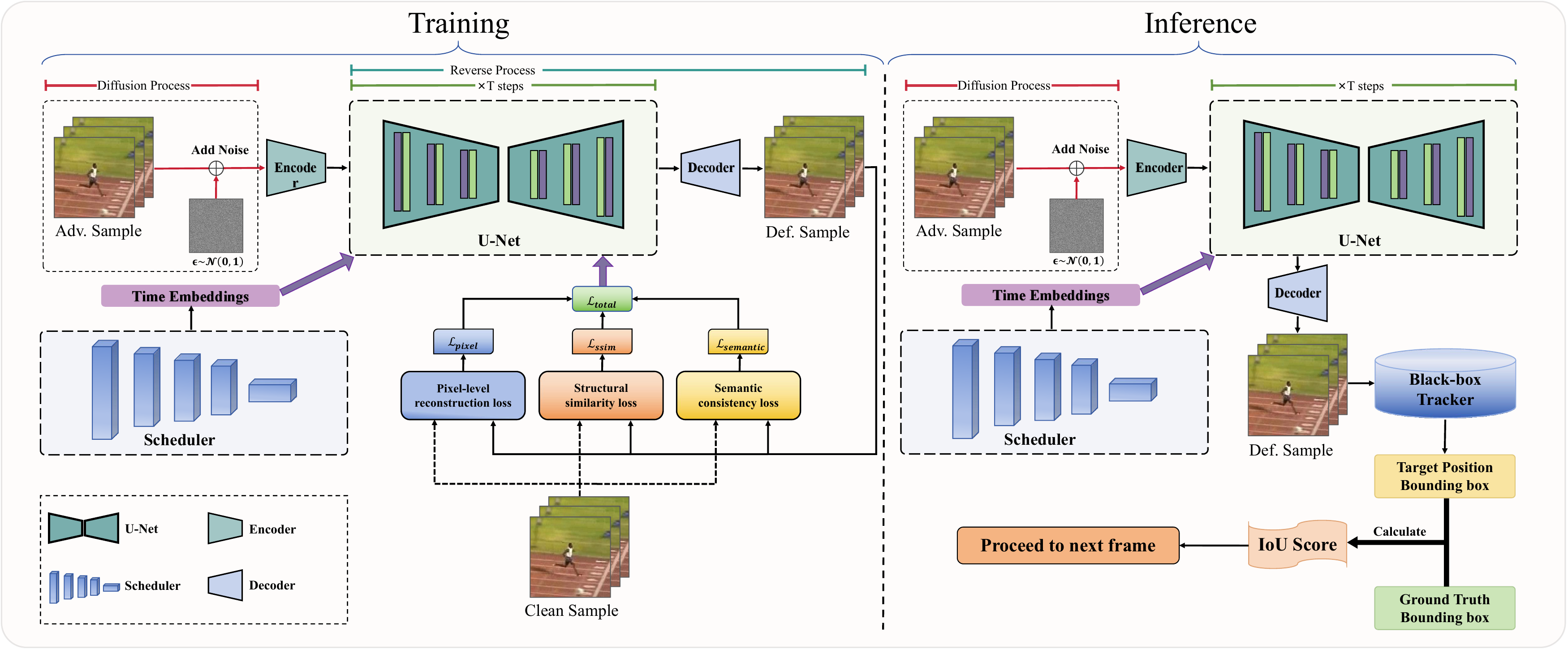}
	\end{center}
	\caption{Overall pipeline of our proposed DDPM-based adversarial defense method, DiffDf.}
	\label{fig:2}
\end{figure}

The overall pipeline of the DiffDf method we proposed is shown in Fig.\ref{fig:2}. In the training phase, images with adversarial perturbations first undergo the forward diffusion process, where Gaussian noise is progressively added to simulate different perturbation levels. These noisy images are then input into a U-Net-based denoising network, where reverse denoising is performed under the guidance of a timestep embedding mechanism and scheduler. To enhance the denoising quality, we introduce three complementary loss functions: pixel-level reconstruction loss for retaining detailed textures, structural similarity loss for improving local structure alignment, and semantic consistency loss, which uses a pre-trained ResNet-50 to align high-level semantic features. These losses collectively form a multi-scale optimization mechanism, effectively enhancing the stability of diffusion recovery and the generalization ability against various types of perturbations. In the inference phase, the adversarial examples processed by DiffDf undergo the same diffusion and denoising process, with the output denoised image directly fed into the black-box tracker, without altering the original architecture. The final output is the target bounding box, which is compared to the ground truth bounding box to compute the IoU score to evaluate robustness, fully demonstrating DiffDf's plug-and-play preprocessing defense capability and versatility.

\section{Experiments}\label{sec:4}

\subsection{Experimental Settings}

We use the GOT-10k\cite{r37} dataset for training, which contains over 10,000 video sequences across more than 500 object categories. The diversity of this dataset ensures that the defense method can adapt to different complex scenarios. Traditional noise modeling typically uses random perturbations such as Gaussian noise and Poisson noise, while the key feature of adversarial examples is that they do not conform to any known random noise distribution. To enable the diffusion model to effectively remove adversarial perturbations, we construct the training data by generating adversarial examples using various adversarial attack methods from visual tracking tasks (\eg white-box CSA attack\cite{r22} and black-box IoU attack\cite{r19}) and pairing them with corresponding clean images, forming one-to-one training pairs. This construction ensures that the defense method learns to remove perturbations and restore the target-specific information when facing different types of adversarial perturbations. We first extract frames from the GOT-10k dataset in the training dataset construction process, and then construct corresponding adversarial examples for each clean image. This allows the model to simultaneously focus on the fidelity of clean images and the recovery of adversarial examples during training. This joint training strategy helps improve the generalization ability of DiffDf, making it highly adaptable to both types of inputs. To enhance data diversity, we uniformly sample every 10th frame from each video sequence to ensure the defense method learns richer scenario variations at different time steps. The training dataset consists of 50,000 pairs of images. The target template is cropped from the initial frame, while the search region is cropped based on the ground-truth annotation in the previous frame, ensuring that the dataset construction reflects real-world scenarios.

\begin{figure*}[t!]
	\begin{center}
		\subfigure{\includegraphics[width=.49\linewidth]{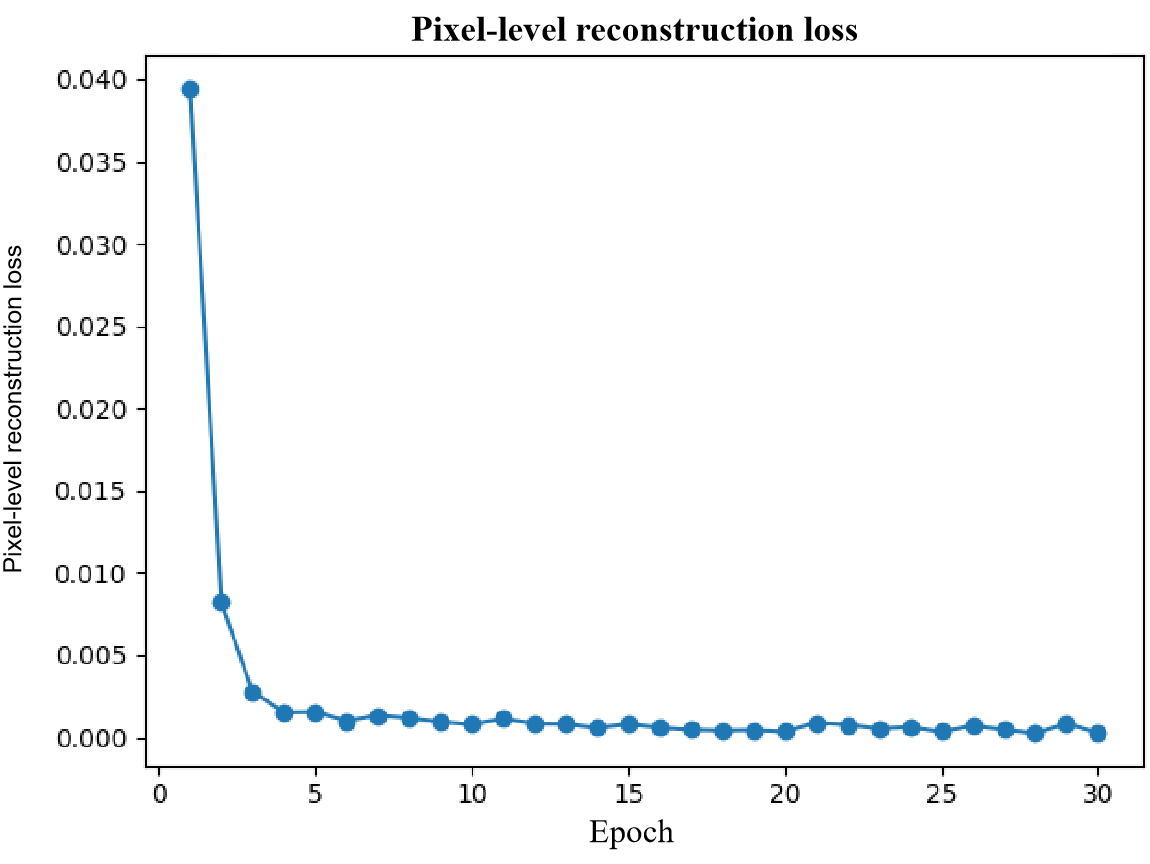}}\hspace{0.5em}
		\subfigure{\includegraphics[width=.49\linewidth]{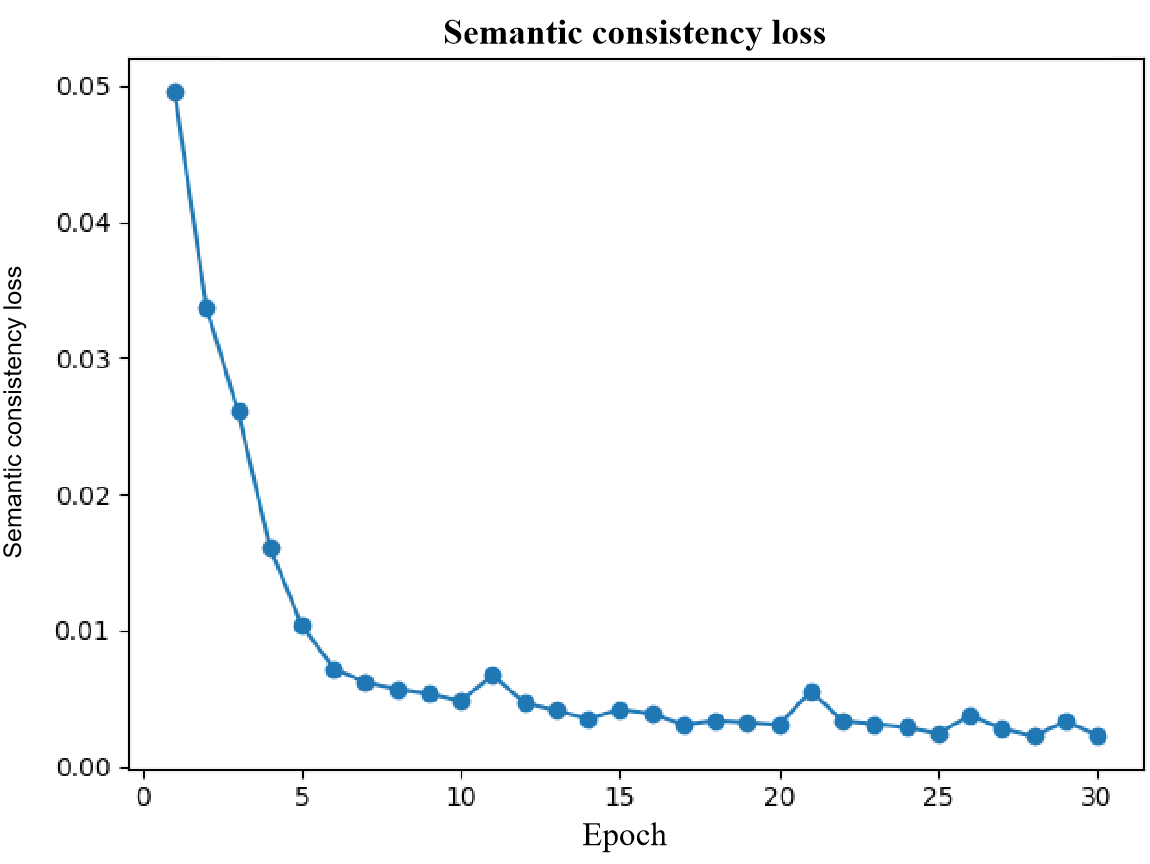}}
		\vfill
		\subfigure{\includegraphics[width=.49\linewidth]{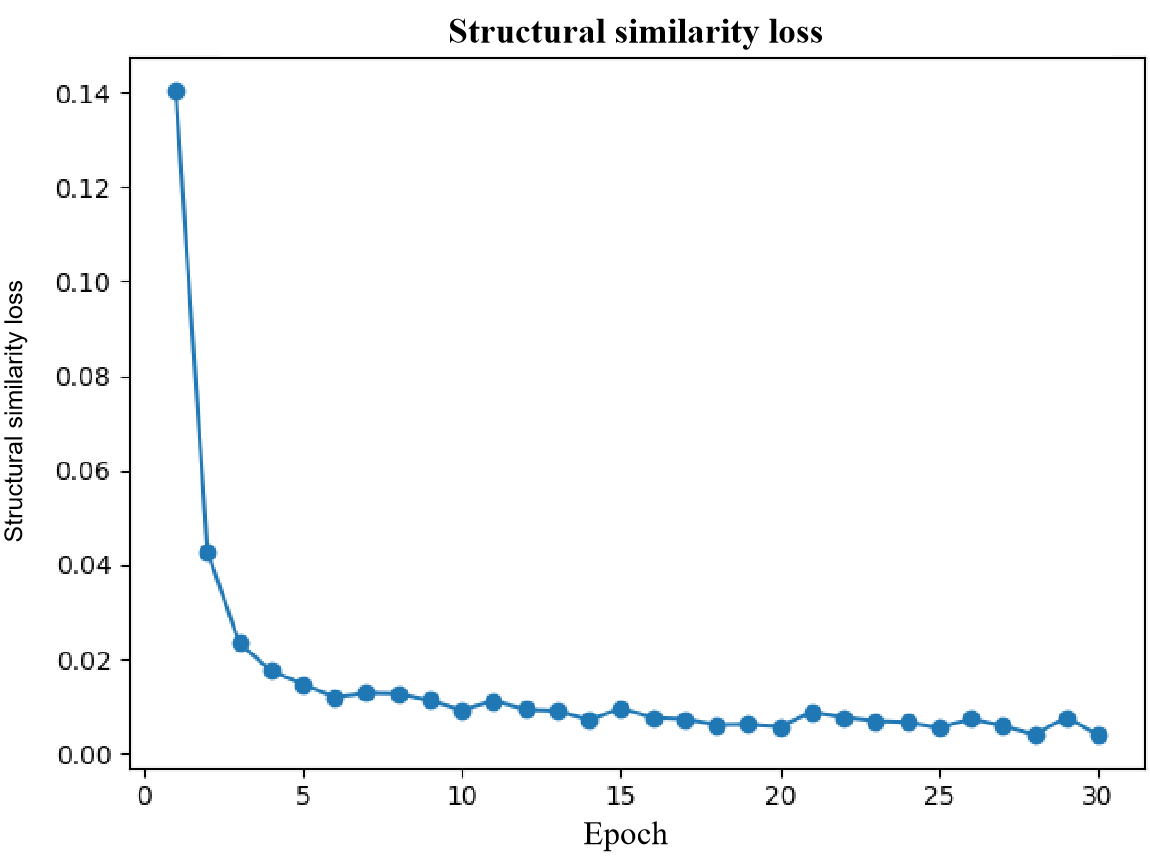}}\hspace{0.5em}
		\subfigure{\includegraphics[width=.49\linewidth]{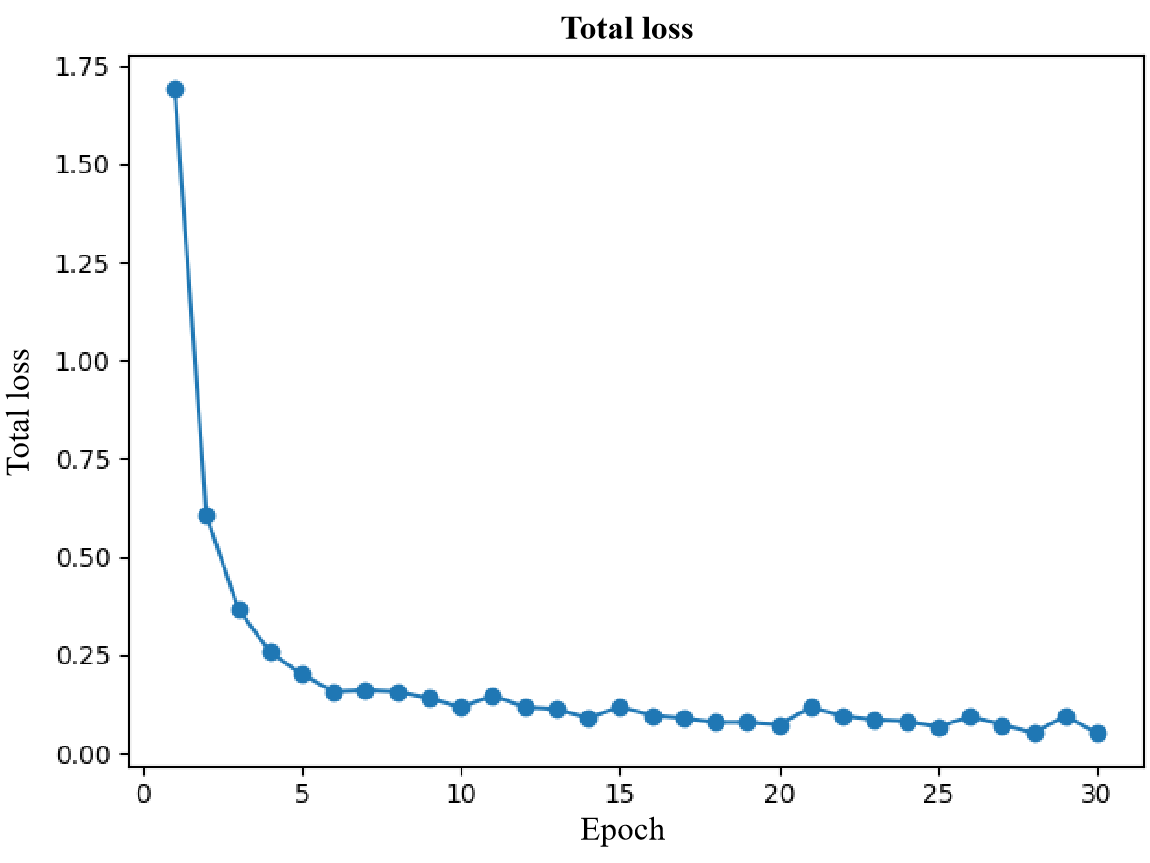}}
	\end{center}
	\caption{The convergence curves of the pixel-level reconstruction loss, semantic consistency loss, structural similarity loss, and the total loss during the training process of our proposed DiffDf method.}
	\label{fig:3}
\end{figure*}

The training process uses the Adam optimizer, with an initial learning rate set to $10^{-4}$, and employs a StepLR scheduler to decay the learning rate by a factor of 10 every 5 epochs. The batch size is set to 8. The hyperparameters $\lambda_1$, $\lambda_2$, and $\lambda_3$ are set to 1, 5, and 10. Our proposed defense method converges after the 6th epoch. The training process takes approximately 13 hours on a server with an AMD EPYC$^{\mathtt{TM}}$ 7542 CPU @ 2.9GHz CPU with 128GB RAM, and an NVIDIA$^\circledR$ GeForce RTX$^{\mathtt{TM}}$ 3090 Ti GPU with 24GB VRAM. Fig.\ref{fig:3} shows the convergence curves of $\mathcal{L}_\text{pixel}$, $\mathcal{L}_\text{semantic}$, and $\mathcal{L}_\text{ssim}$ (to illustrate clearly, we use the curve of 1-$\mathcal{L}_\text{ssim}$ instead), as well as $\mathcal{L}_\text{total}$ during training. All four loss curves dramatically decrease and converge within the first 6 epochs. In the later stages, the loss values remain low, further validating the effectiveness and stability of the diffusion model. By jointly imposing multi-scale constraints at the pixel, semantic, and structural levels, the method can effectively remove adversarial perturbations and restore the target-specific information.

We validate the performance of the proposed DiffDf method on multiple popular tracking datasets, including the short-term tracking dataset VOT2018\cite{r38}, the mid-term tracking dataset OTB2015\cite{r39}, and the long-term tracking dataset LaSOT\cite{r40}. The VOT2018 dataset contains 60 challenging short-term tracking video sequences. The evaluation metrics include \textit{Accuracy}, \textit{Robustness}, \textit{Lost number}, and \textit{Expected Average Overlap} (EAO). Accuracy measures the average overlap rate between the predicted and ground truth bounding boxes; Robustness and Lost number assess the frequency of tracker failure; EAO combines both accuracy and robustness. OTB2015 is a classic tracking dataset consisting of 100 mid-term video sequences. The evaluation metrics include \textit{Success} and \textit{Precision}. Success refers to the proportion of video frames where the average overlap rate between the predicted and ground truth bounding boxes exceeds a predefined threshold (usually 0.5); Precision measures the proportion of frames where the Euclidean distance between the predicted and ground truth bounding box centers is less than a threshold (typically 20 pixels). The LaSOT dataset is a large-scale tracking dataset containing over 1,400 long-term video sequences. The evaluation metrics include \textit{Success} and \textit{Normalized Precision} (P$_{\text{norm}}$). Success is defined the same as in OTB2015, while P$_{\text{norm}}$ normalizes Precision as defined in OTB2015 based on the target scale, making it more suitable for long-term tracking scenarios.

\subsection{Comparison results}

\begin{table}[h]
	\centering
	\caption{Results of our proposed DiffDf method on defending the SiamRPN++ tracker\cite{r2} against the white-box CSA attack\cite{r22} on the VOT2018 dataset\cite{r38}.}
	\label{tab:1}
		\begin{tabular}{l|cccc}
			\toprule
			Methods       & Accuracy $\uparrow$ & Robustness $\downarrow$ & Lost number $\downarrow$ & EAO $\uparrow$   \\
			\midrule
			\rowcolor{lightgray!40}
			Original     & 0.609    & 0.276      & 59          & 0.374 \\
			CSA-S        & 0.508    & 1.456      & 311         & 0.090  \\
			\rowcolor{yellow!40}
			CSA-S+DiffDf  & 0.577    & 0.295      & 63          & 0.332 \\
			CSA-T         & 0.541    & 1.147      & 245         & 0.123 \\
			\rowcolor{yellow!40}
			CSA-T+DiffDf  & 0.596    & 0.300      & 64          & 0.341 \\
			CSA-TS        & 0.467    & 2.013      & 430         & 0.073 \\
			\rowcolor{yellow!40}
			CSA-TS+DiffDf & 0.578    & 0.337      & 72          & 0.305 \\
			\bottomrule
		\end{tabular}%
\end{table}

\begin{table}[h]
	\centering
	\caption{Results of our proposed DiffDf method on defending the SiamRPN++ tracker\cite{r2} against the white-box CSA attack\cite{r22} on the OTB2015\cite{r39} and LaSOT\cite{r40} datasets.}
	\label{tab:2}
		\begin{tabular}{l|ccccc}
			\toprule
			\multirow{2}{*}{Methods} & \multicolumn{2}{c}{OTB2015} &  & \multicolumn{2}{c}{LaSOT} \\ \cline{2-6} 
			& Success $\uparrow$     & Precision $\uparrow$    &  & Success $\uparrow$     & P$_{\text{norm}} \uparrow$    \\
			\midrule
			\rowcolor{lightgray!40}
			Original   & 0.697       & 0.917        &  & 0.496       & 0.569       \\
			CSA-S                    & 0.346       & 0.486        &  & 0.180       & 0.219       \\
			\rowcolor{yellow!40}
			CSA-S+DiffDf             & 0.658       & 0.857        &  & 0.484       & 0.557       \\
			CSA-T                    & 0.527       & 0.715        &  & 0.393       & 0.448       \\
			\rowcolor{yellow!40}
			CSA-T+DiffDf             & 0.665       & 0.871        &  & 0.492       & 0.565       \\
			CSA-TS                   & 0.322       & 0.467        &  & 0.168       & 0.201       \\
			\rowcolor{yellow!40}
			CSA-TS+DiffDf            & 0.656       & 0.848        &  & 0.466       & 0.541    \\
			\bottomrule
		\end{tabular}%
\end{table}

To comprehensively evaluate the robustness of the proposed DiffDf defense method across different benchmark datasets, we conducted experimental comparisons using the white-box CSA attack method\cite{r22} and the SiamRPN++\cite{r2} tracker on three mainstream datasets: VOT2018\cite{r38}, OTB2015\cite{r39}, and LaSOT\cite{r40}. The results are presented in Table \ref{tab:1} and Table \ref{tab:2}. Specifically, for the Siamese network architecture adopted by SiamRPN++, we apply the white-box CSA attack separately to the target template, the search region, and both simultaneously, denoted as CSA-T, CSA-S, and CSA-TS. Correspondingly, the results after applying our DiffDf defense method are denoted as CSA-T+DiffDf, CSA-S+DiffDf, and CSA-TS+DiffDf.

In the short-term tracking scenario of VOT2018\cite{r38}, the white-box CSA attack on the search region (CSA-S) severely impacted the performance of SiamRPN++, with the number of target losses increasing dramatically from 59 to 443, and the EAO dropping from 0.374 to 0.073 with a decrease of 80.5\%. However, after applying the proposed DiffDf defense (CSA-S+DiffDf), the number of target losses significantly decreased to 63, and the EAO recovered to 0.332, showing a recovery rate of 89.3\%. Notably, the defense performance against the white-box CSA attack on the target template (CSA-T) was even more prominent. After applying DiffDf (CSA-T+DiffDf), the EAO rose to 0.341, recovering to 91.2\% of the baseline of 0.374.

\begin{figure}[t!]
	\begin{center}
		\subfigure{\includegraphics[width=.328\linewidth]{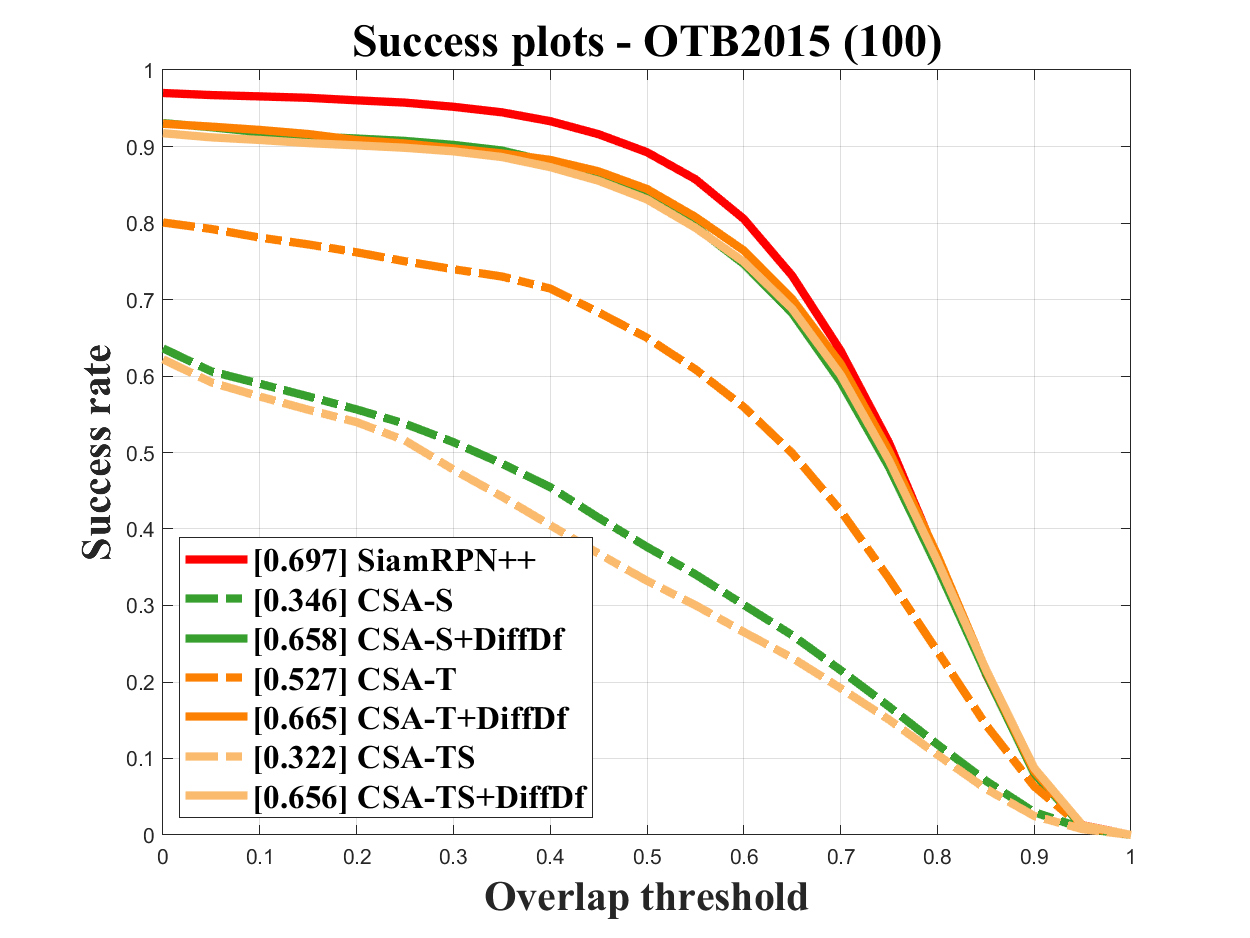}}
		\hspace{0.05em}
		\subfigure{\includegraphics[width=.328\linewidth]{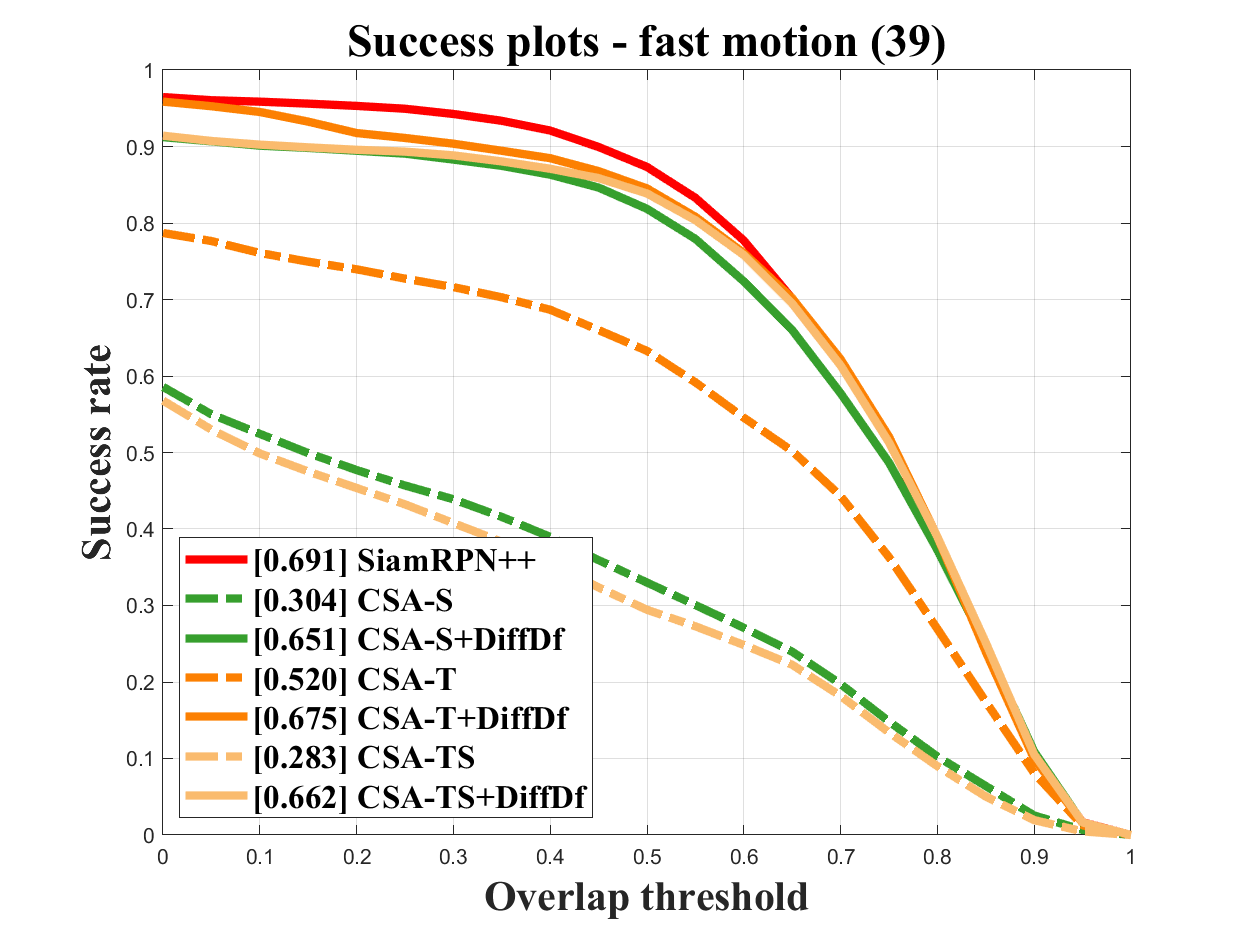}}
		\hspace{0.05em}
		\subfigure{\includegraphics[width=.328\linewidth]{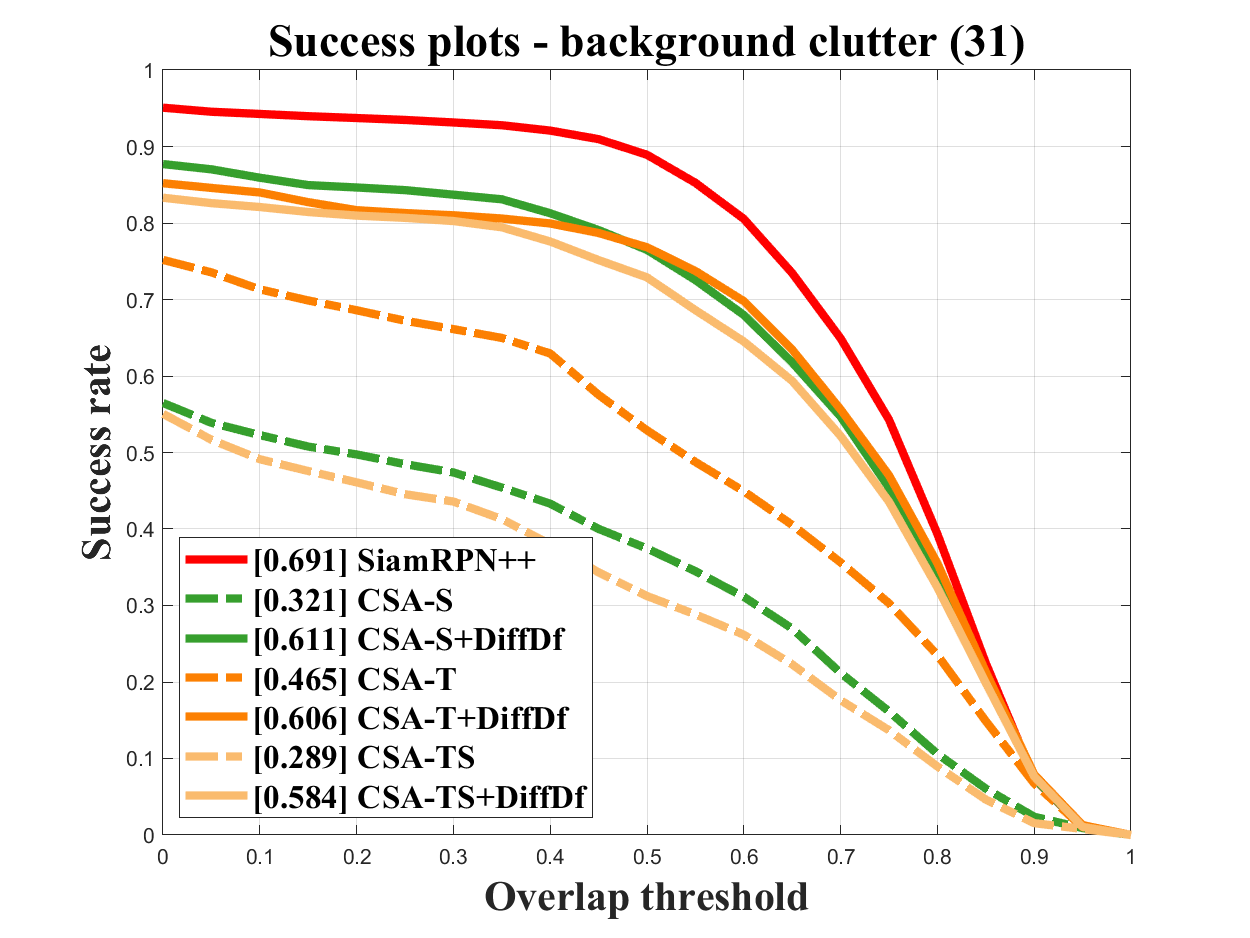}}
		\vfill
		\subfigure{\includegraphics[width=.328\linewidth]{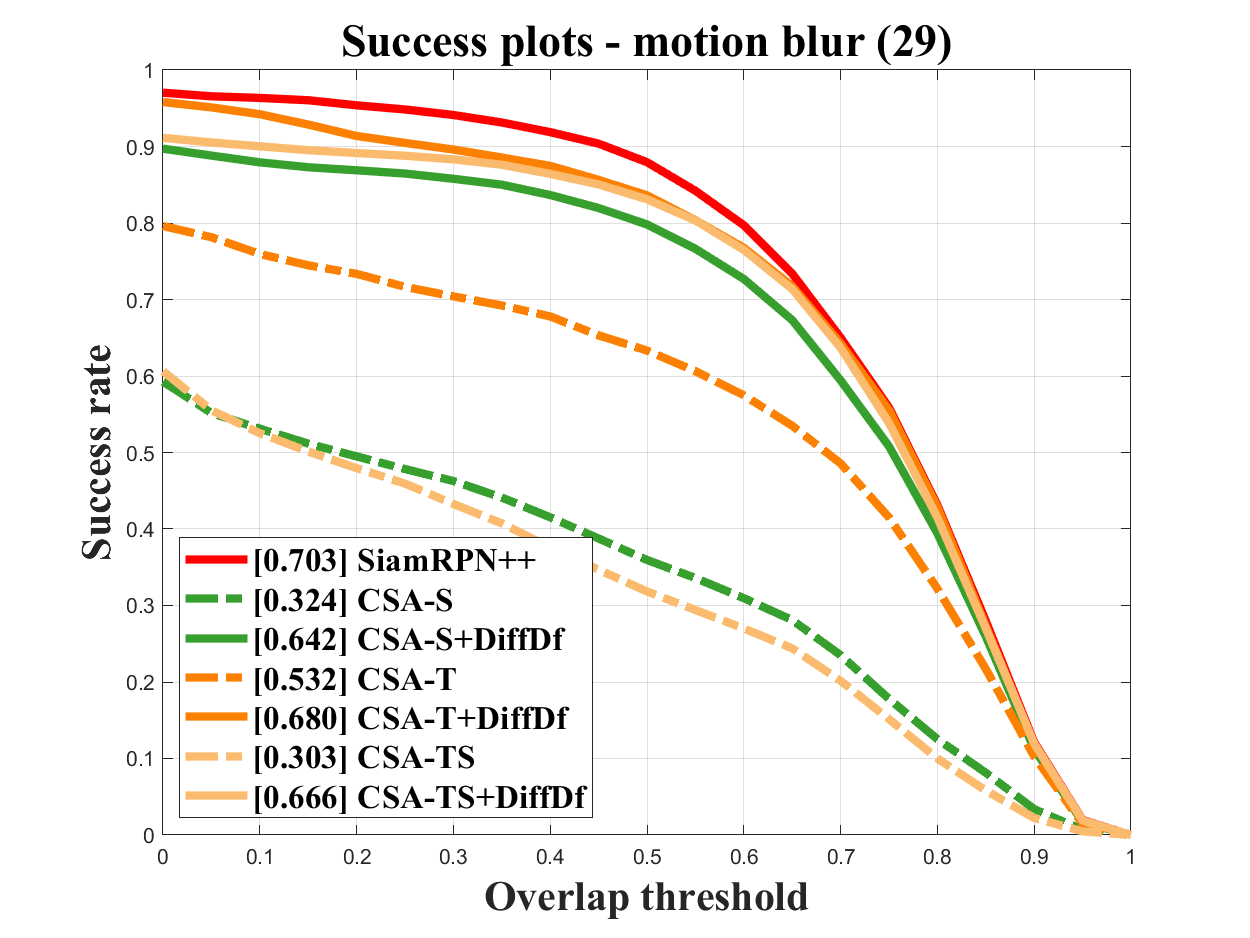}}
		\hspace{0.05em}
		\subfigure{\includegraphics[width=.328\linewidth]{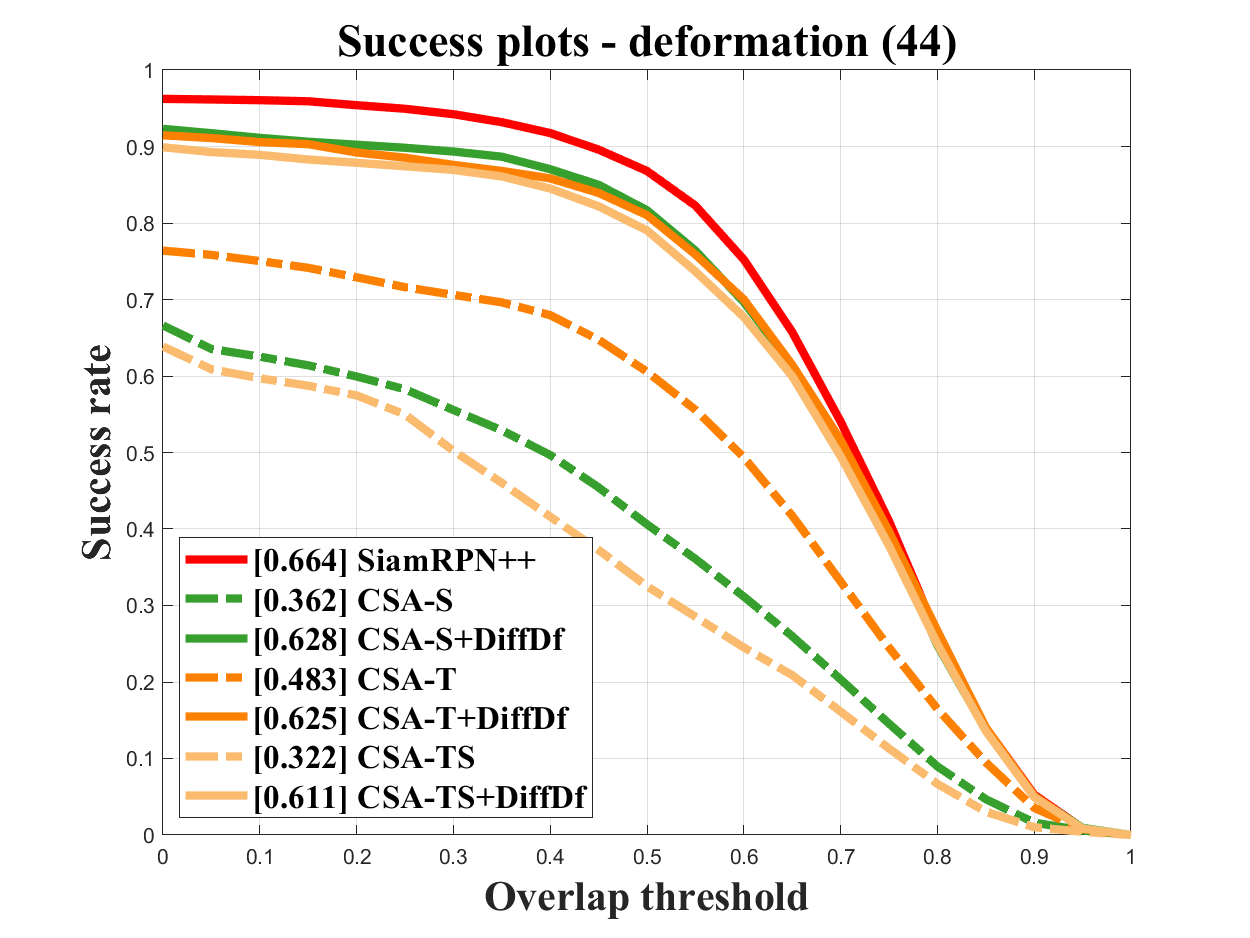}}
		\hspace{0.05em}
		\subfigure{\includegraphics[width=.328\linewidth]{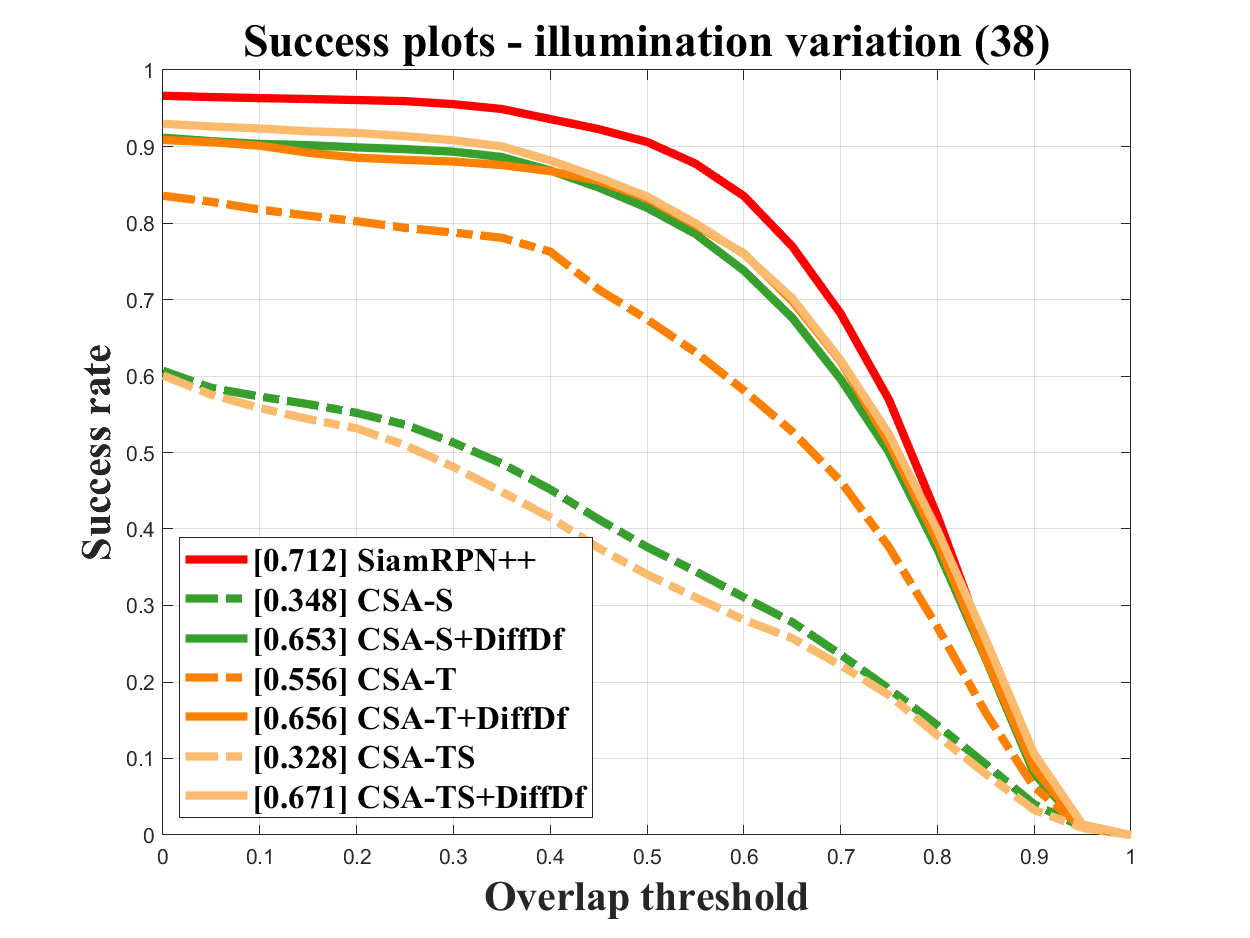}}
		\vfill
		\subfigure{\includegraphics[width=.328\linewidth]{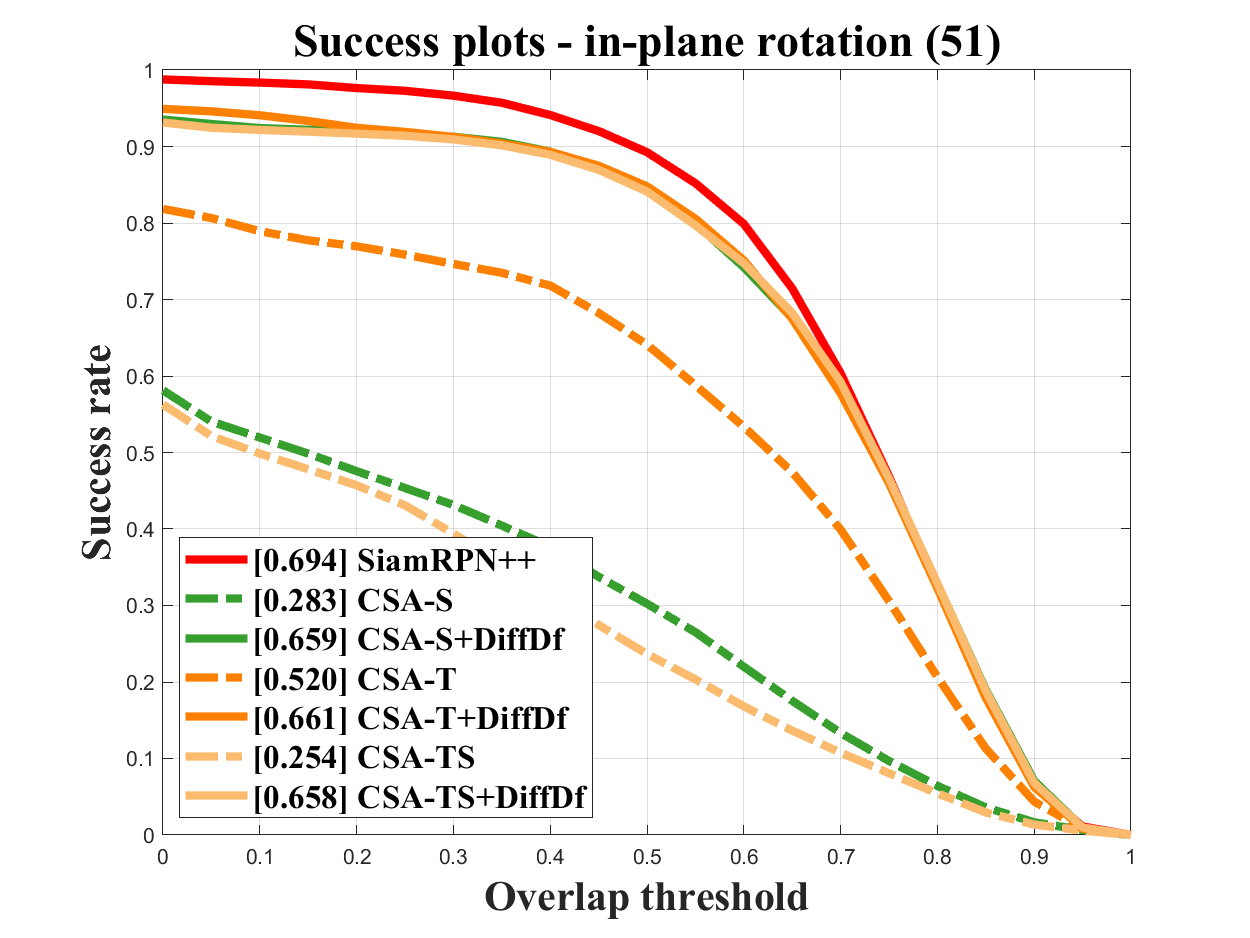}}
		\hspace{0.05em}
		\subfigure{\includegraphics[width=.328\linewidth]{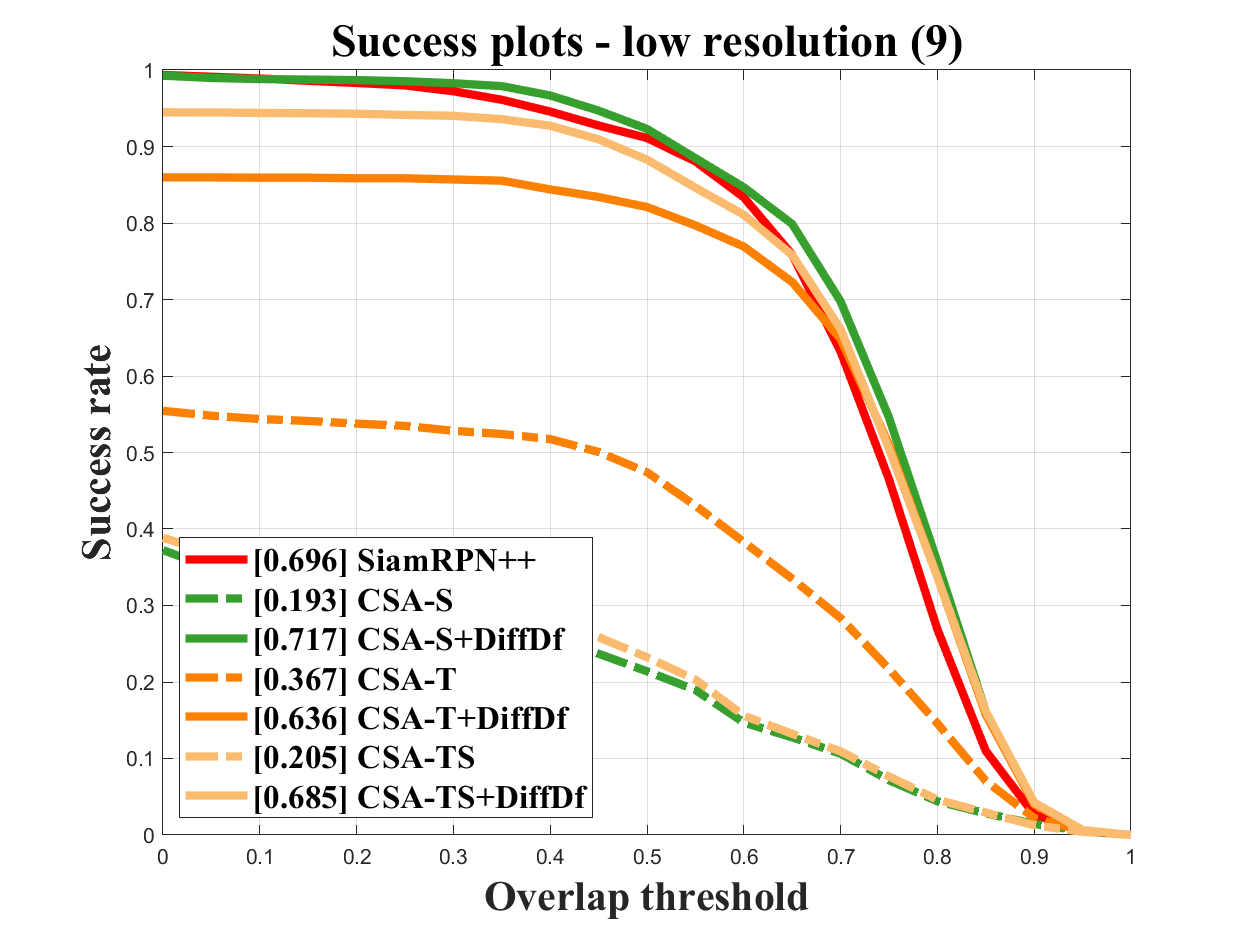}}
		\hspace{0.05em}
		\subfigure{\includegraphics[width=.328\linewidth]{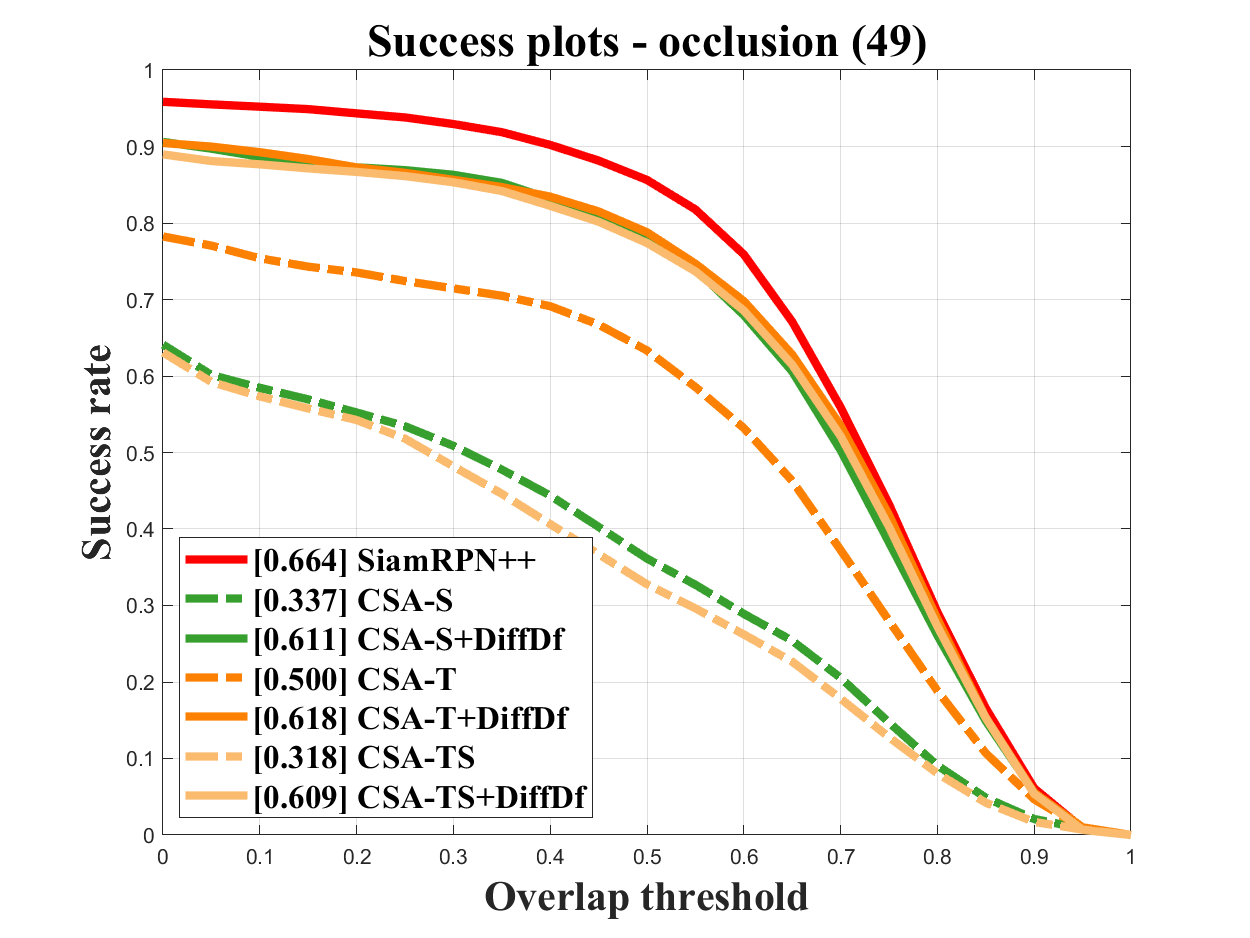}}
		\vfill
		\subfigure{\includegraphics[width=.328\linewidth]{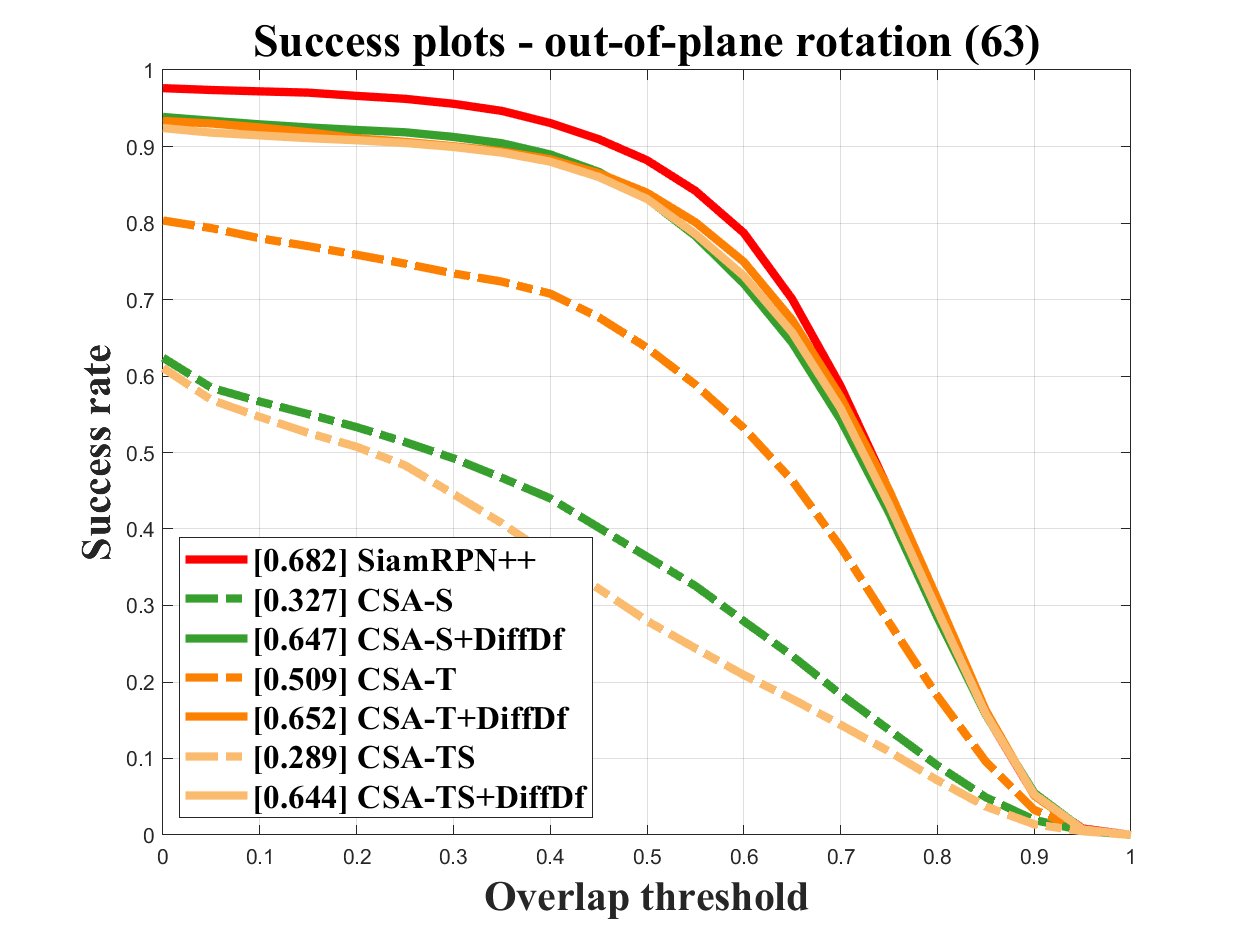}}
		\hspace{0.05em}
		\subfigure{\includegraphics[width=.328\linewidth]{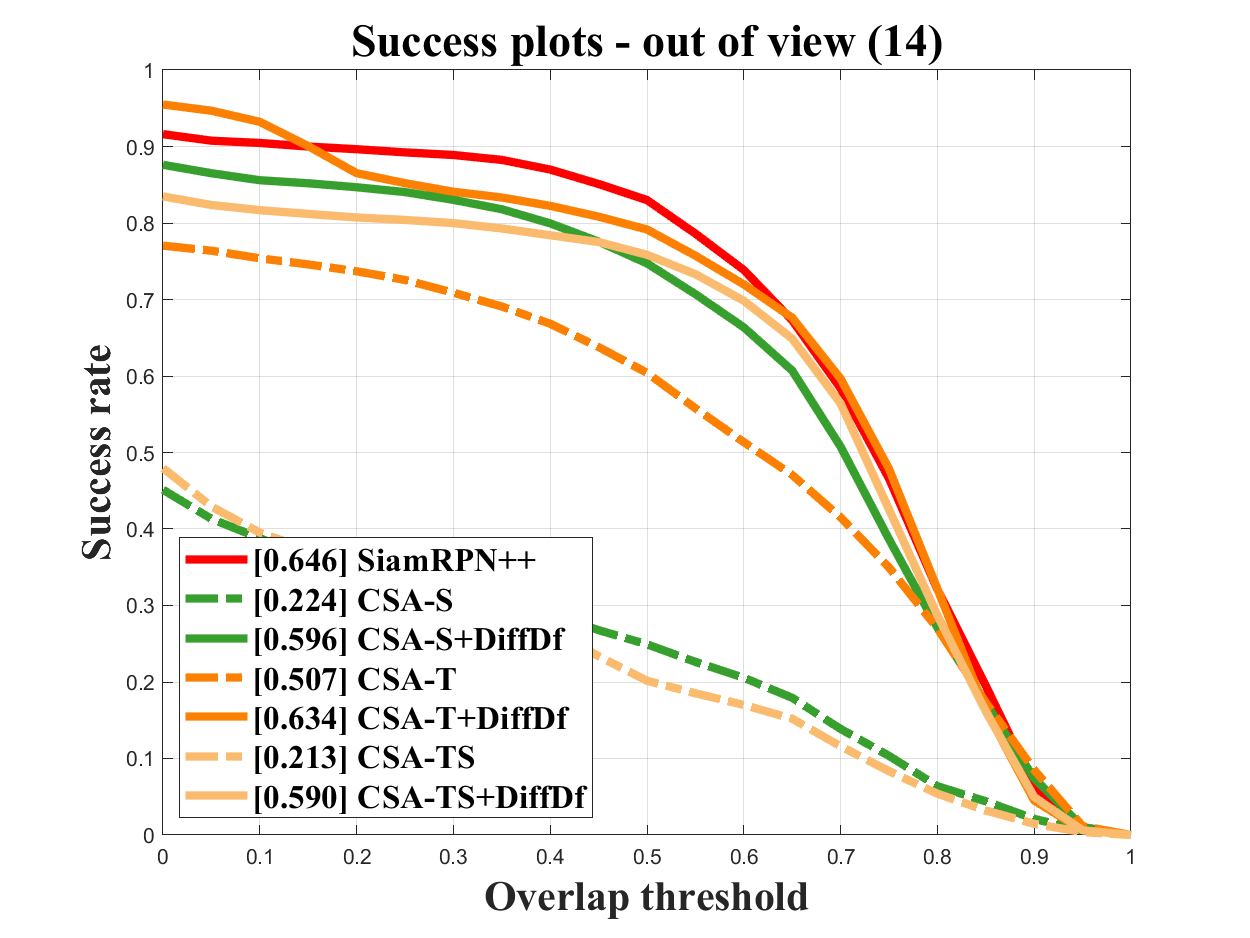}}
		\hspace{0.05em}
		\subfigure{\includegraphics[width=.328\linewidth]{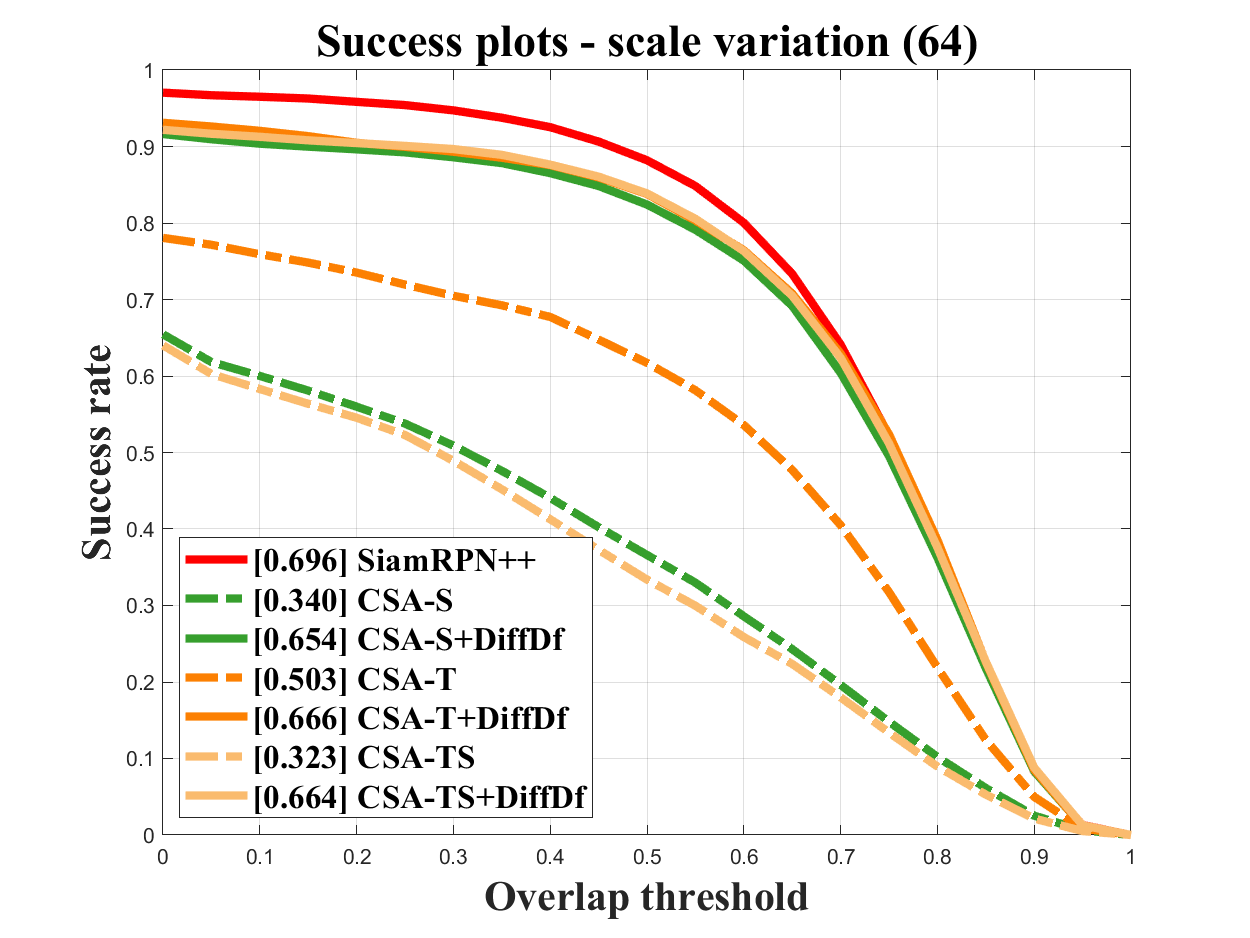}}
	\end{center}
	\caption{Success plots of our proposed DiffDf method on defending the SiamRPN++ tracker\cite{r2} against the white-box CSA attacks\cite{r22} on the OTB2015 dataset\cite{r39}.}
	\label{fig:4}
\end{figure}

\begin{figure}[t!]
	\begin{center}
		\subfigure{\includegraphics[width=.328\linewidth]{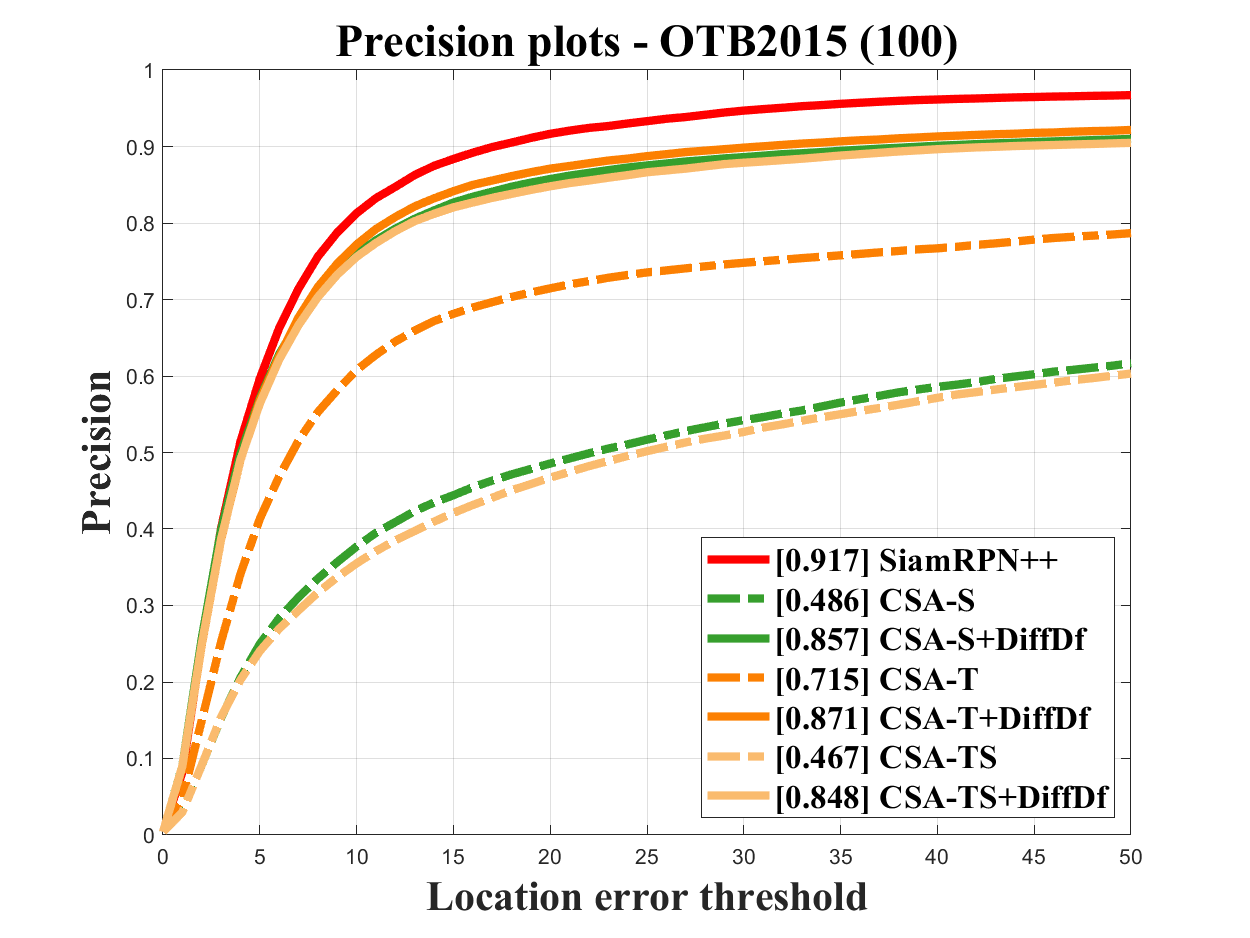}}
		\hspace{0.05em}
		\subfigure{\includegraphics[width=.328\linewidth]{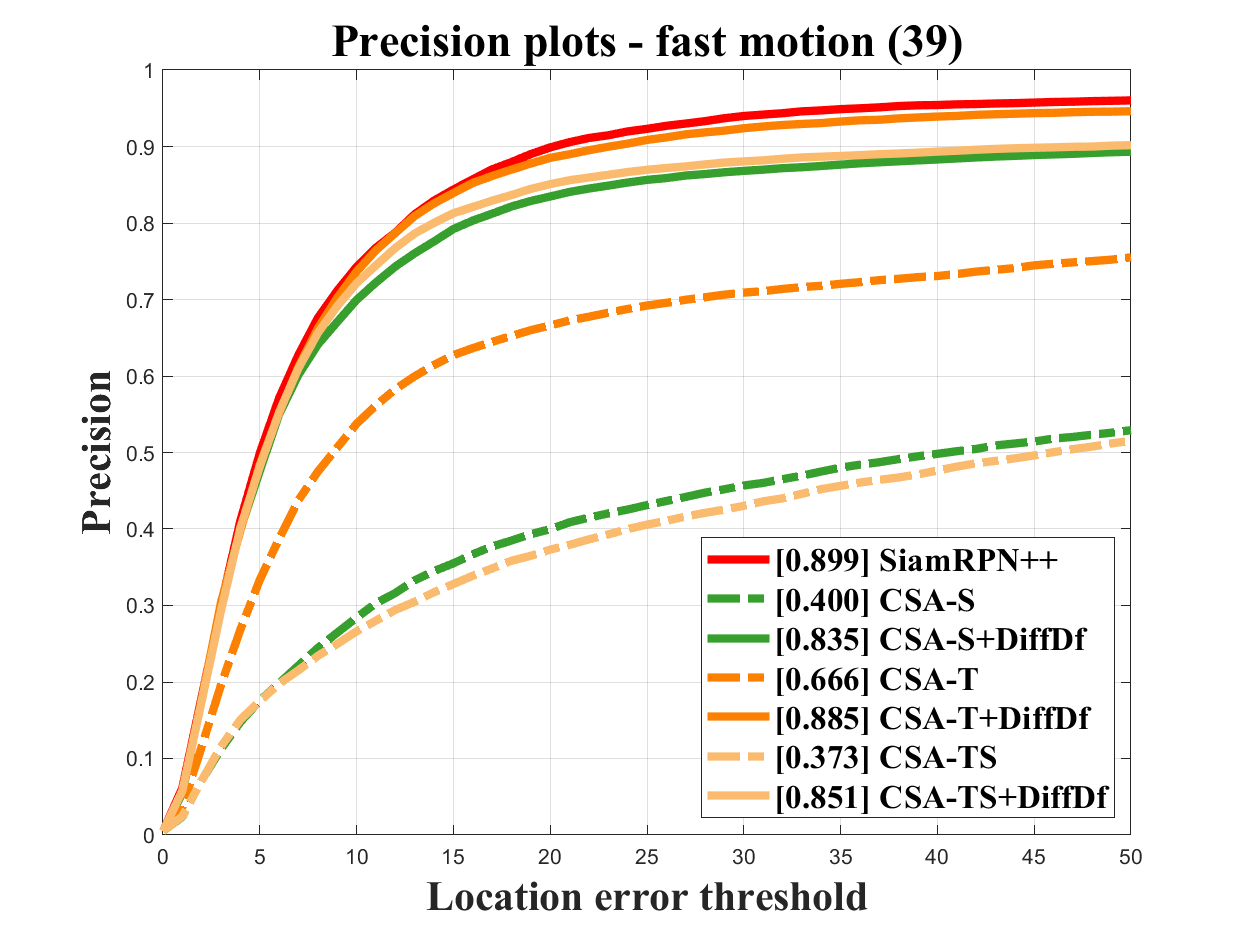}}
		\hspace{0.05em}
		\subfigure{\includegraphics[width=.328\linewidth]{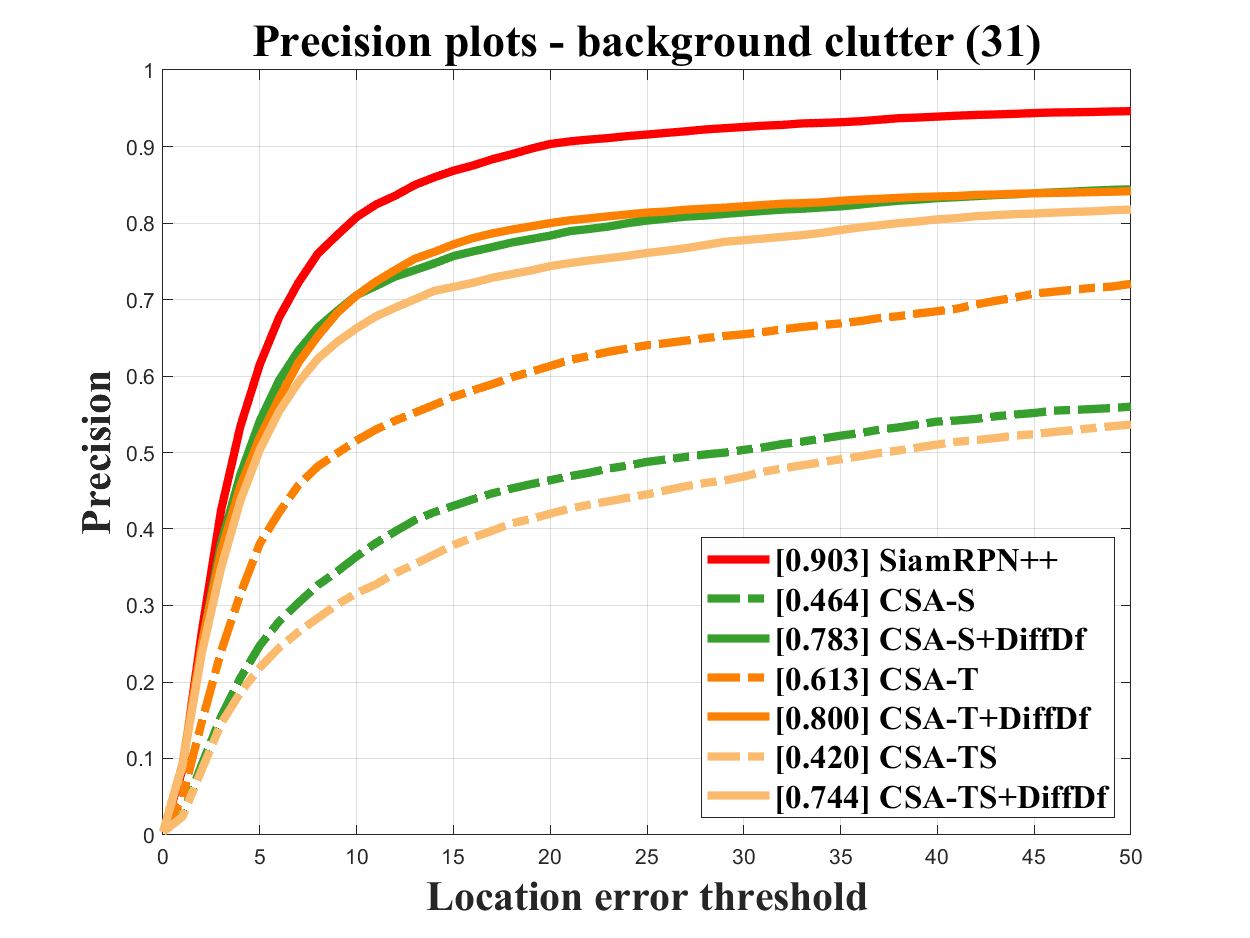}}
		\vfill
		\subfigure{\includegraphics[width=.328\linewidth]{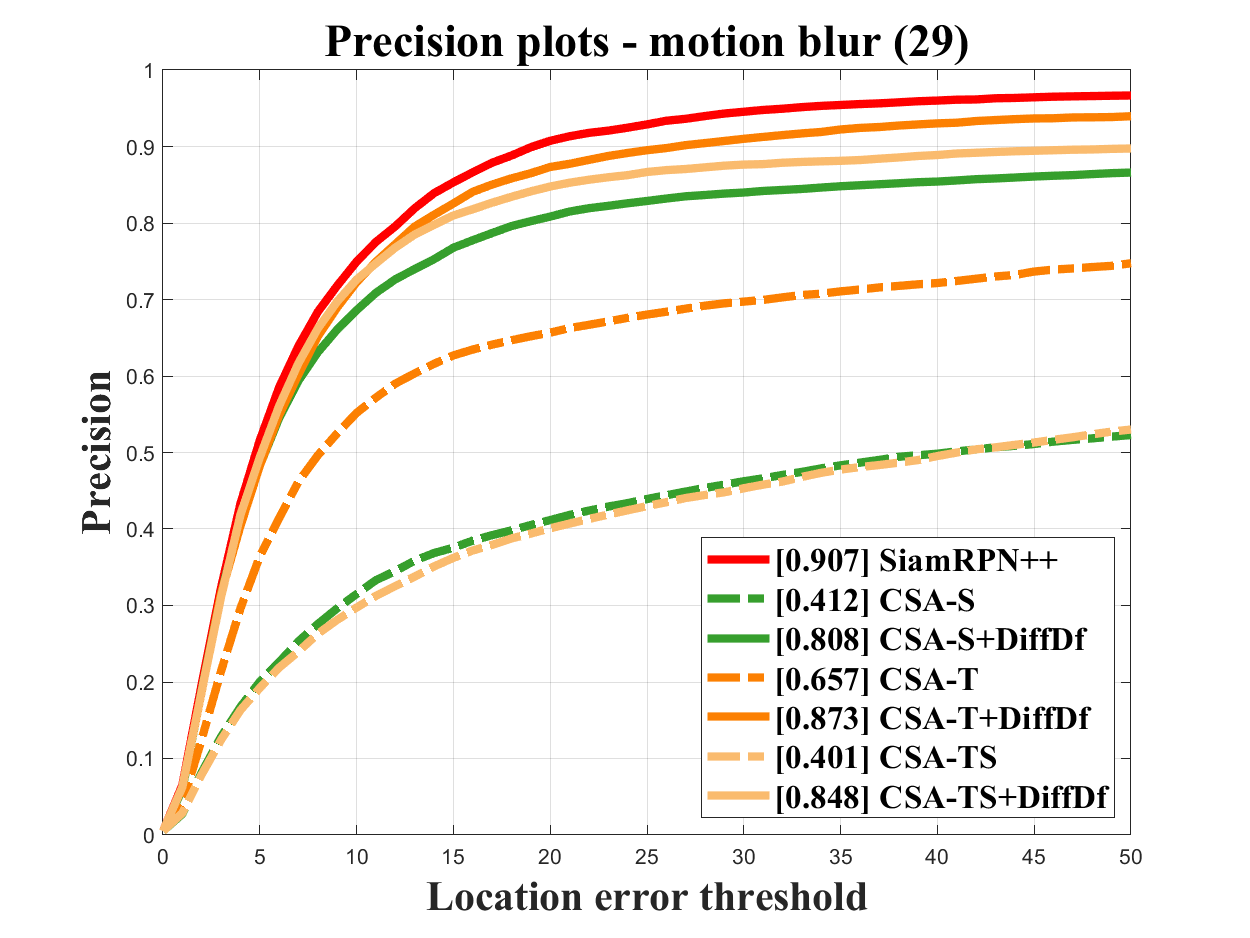}}
		\hspace{0.05em}
		\subfigure{\includegraphics[width=.328\linewidth]{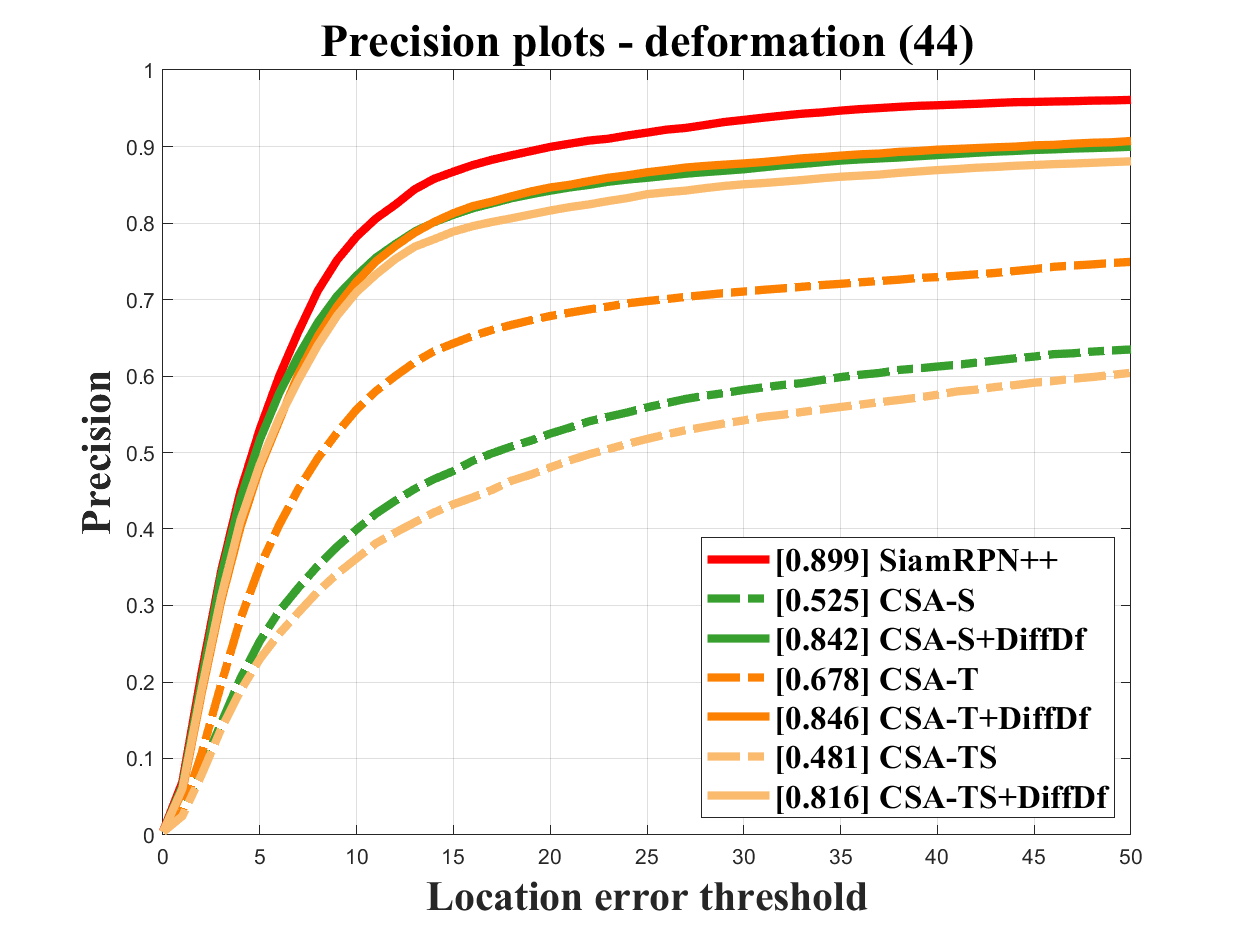}}
		\hspace{0.05em}
		\subfigure{\includegraphics[width=.328\linewidth]{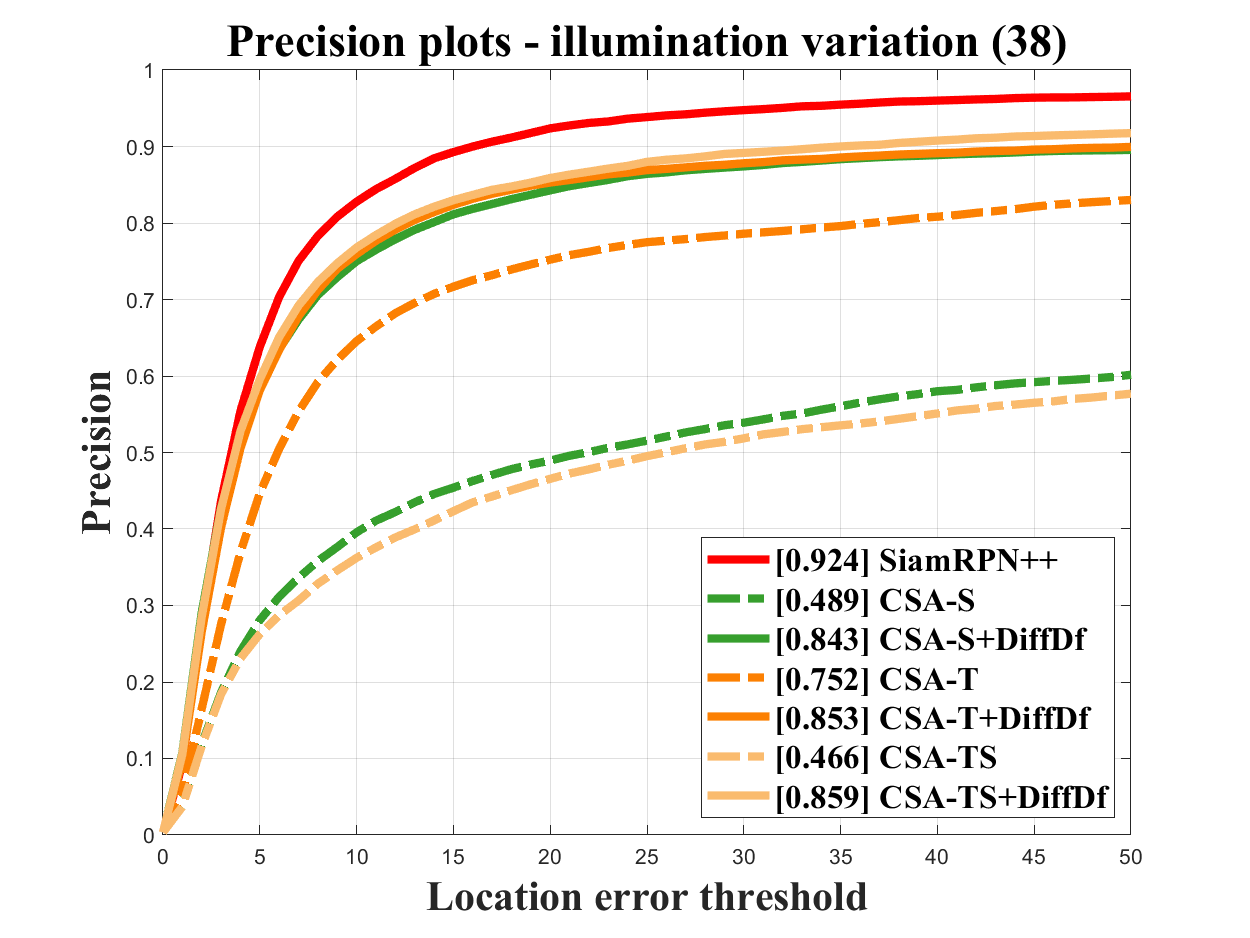}}
		\vfill
		\subfigure{\includegraphics[width=.328\linewidth]{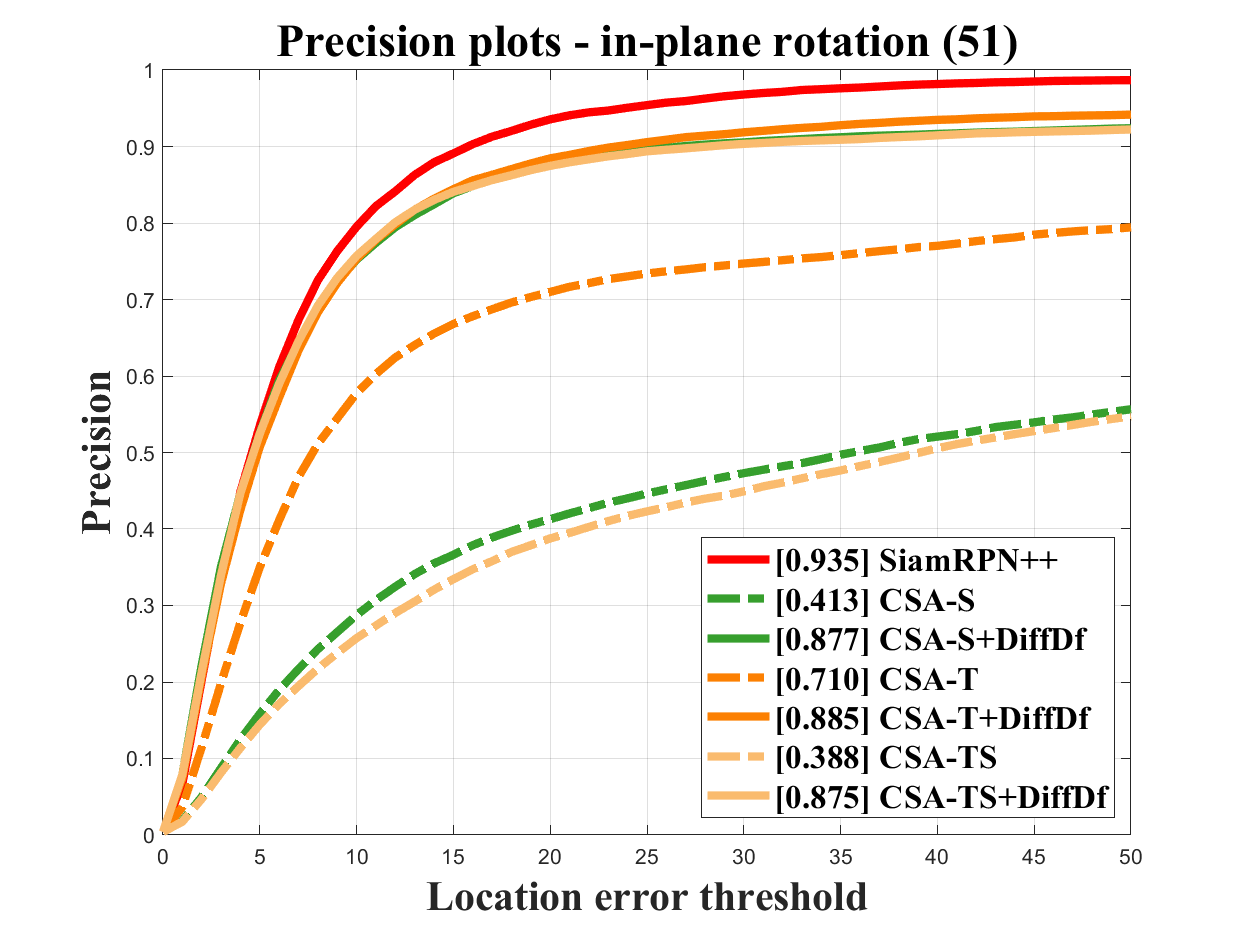}}
		\hspace{0.05em}
		\subfigure{\includegraphics[width=.328\linewidth]{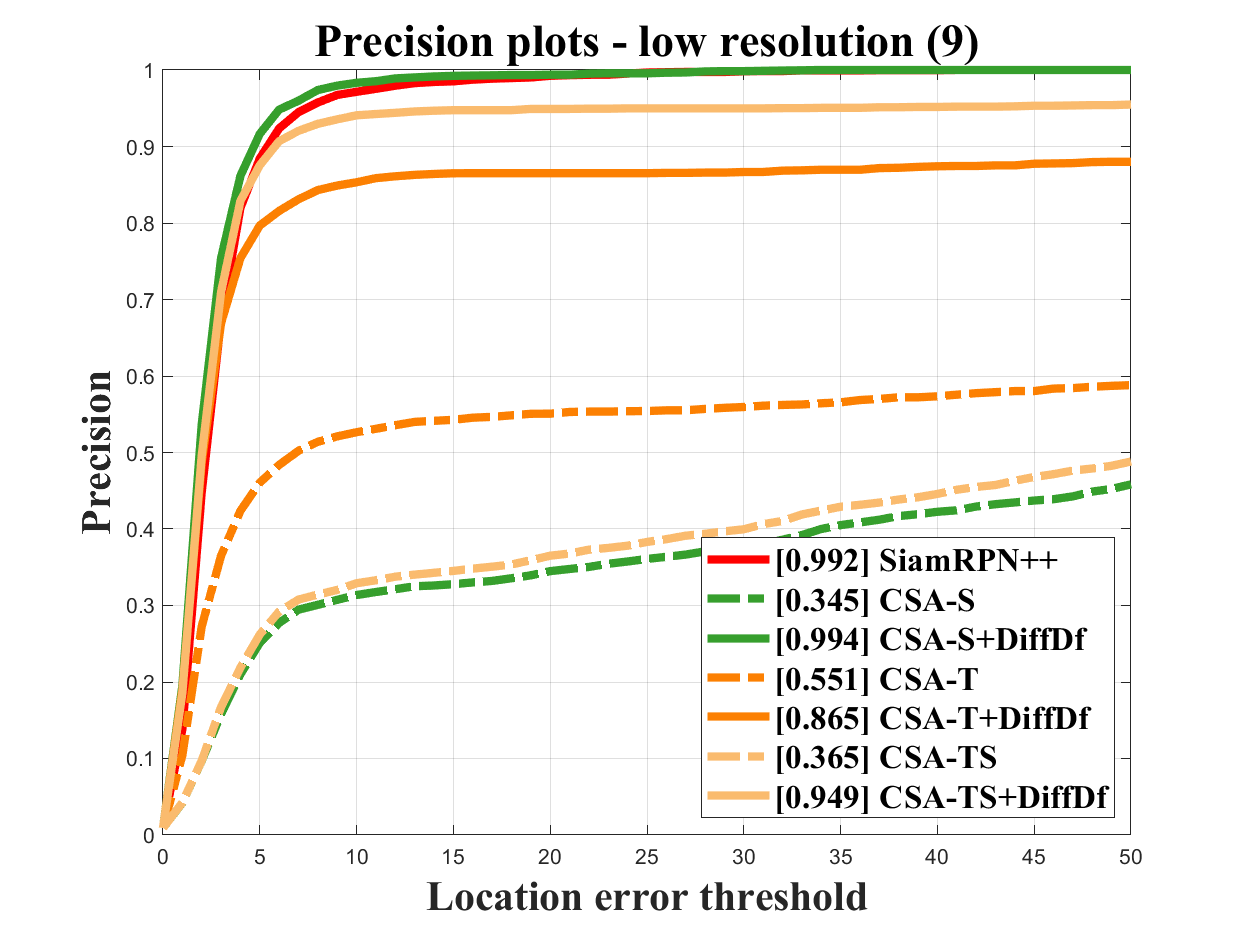}}
		\hspace{0.05em}
		\subfigure{\includegraphics[width=.328\linewidth]{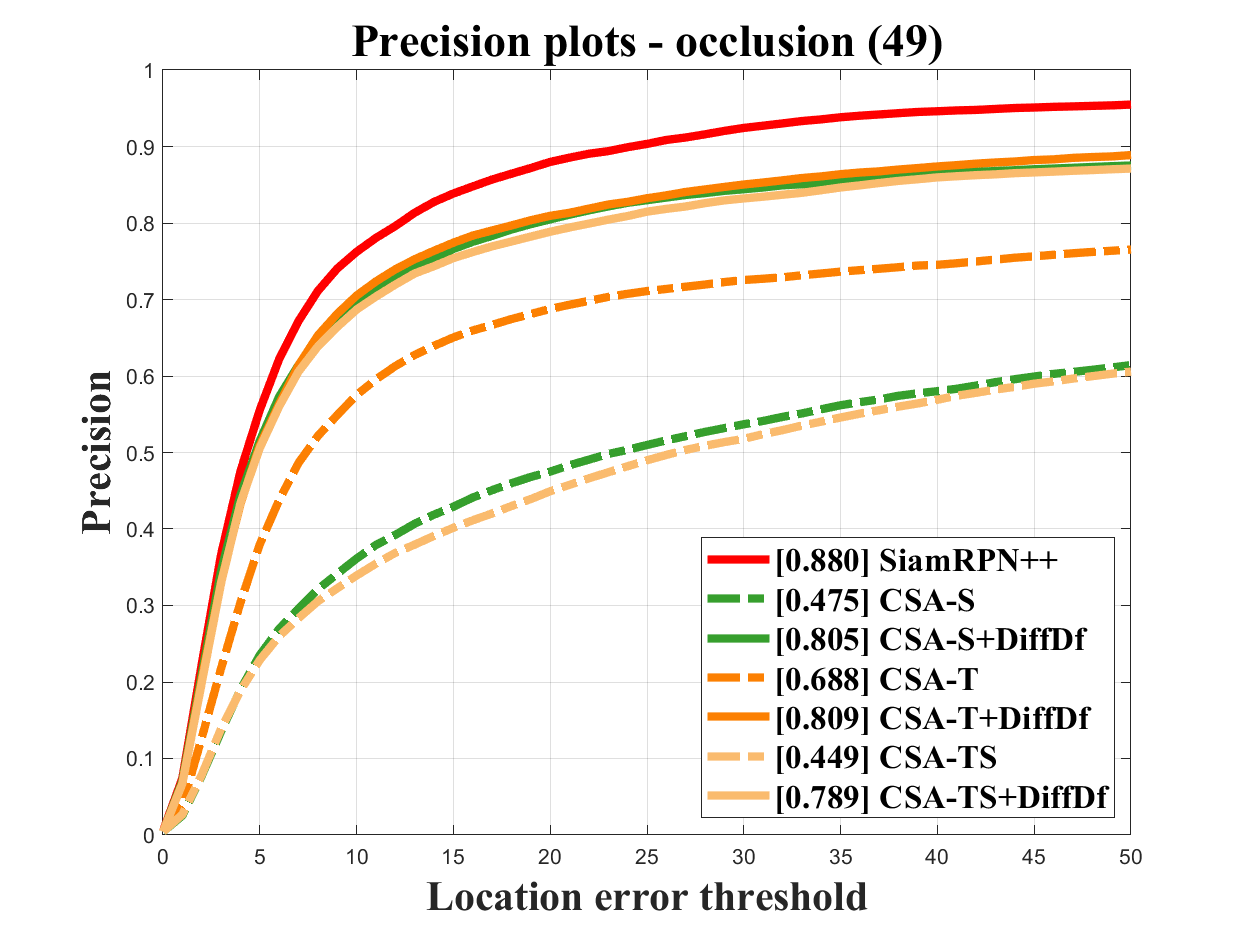}}
		\vfill
		\subfigure{\includegraphics[width=.328\linewidth]{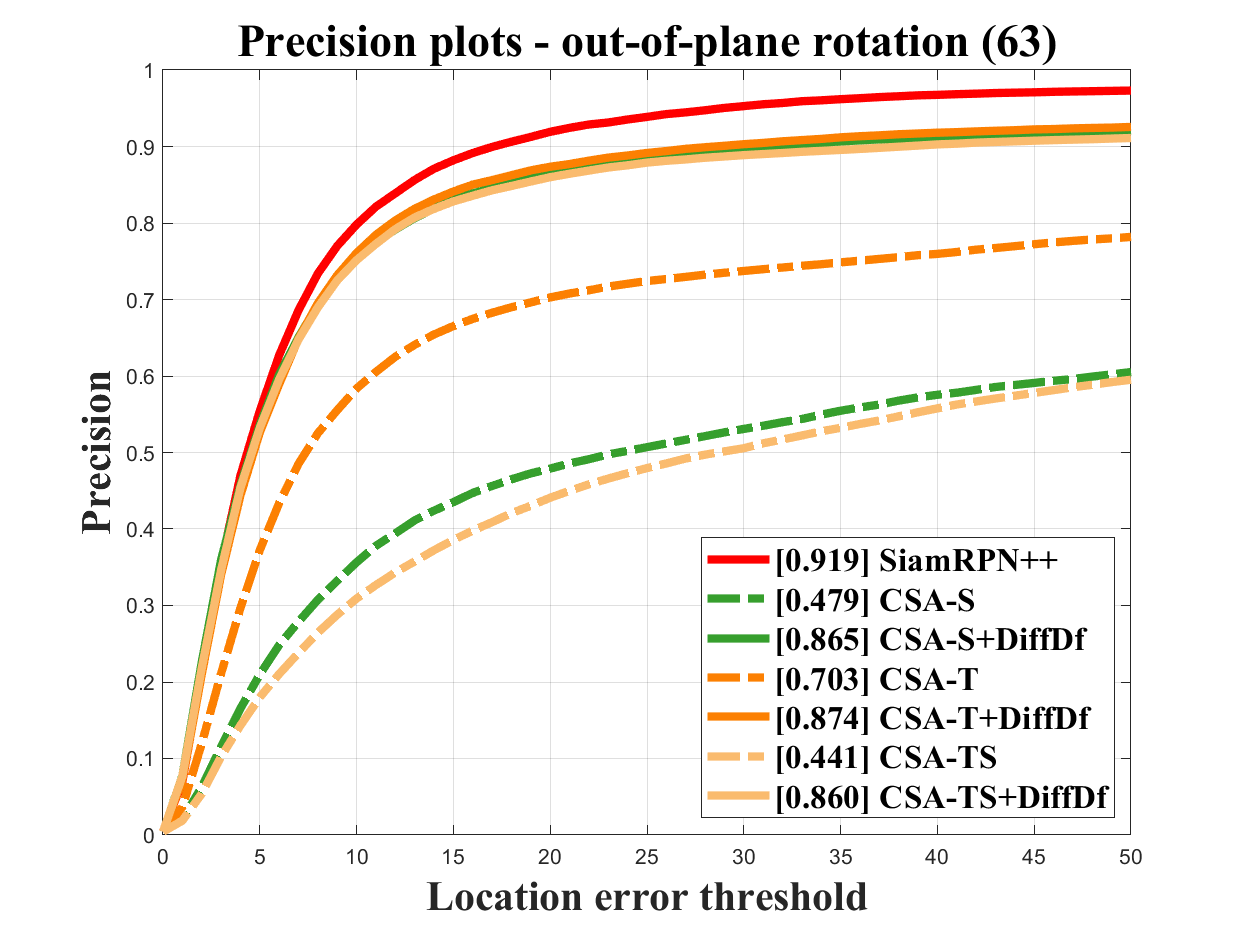}}
		\hspace{0.05em}
		\subfigure{\includegraphics[width=.328\linewidth]{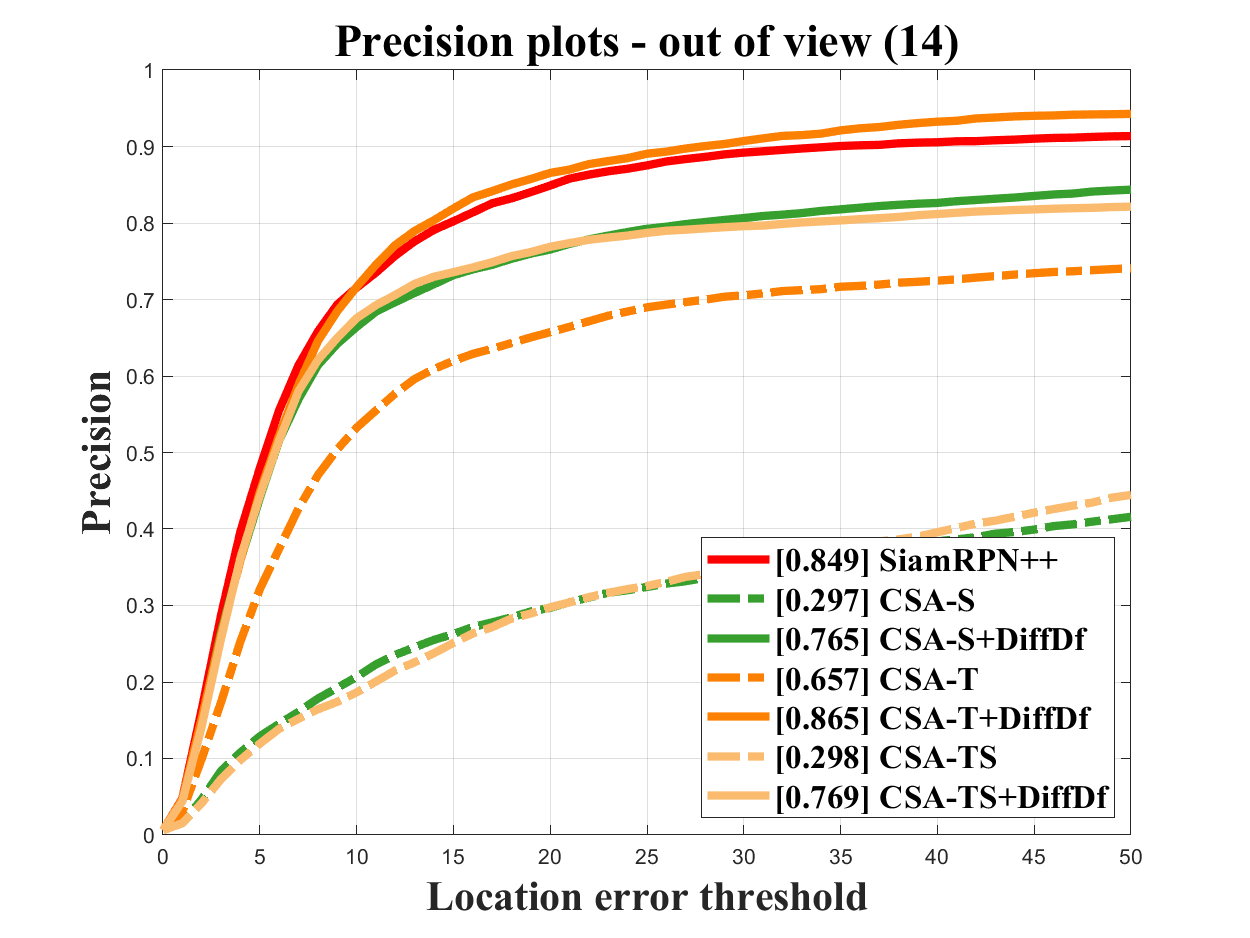}}
		\hspace{0.05em}
		\subfigure{\includegraphics[width=.328\linewidth]{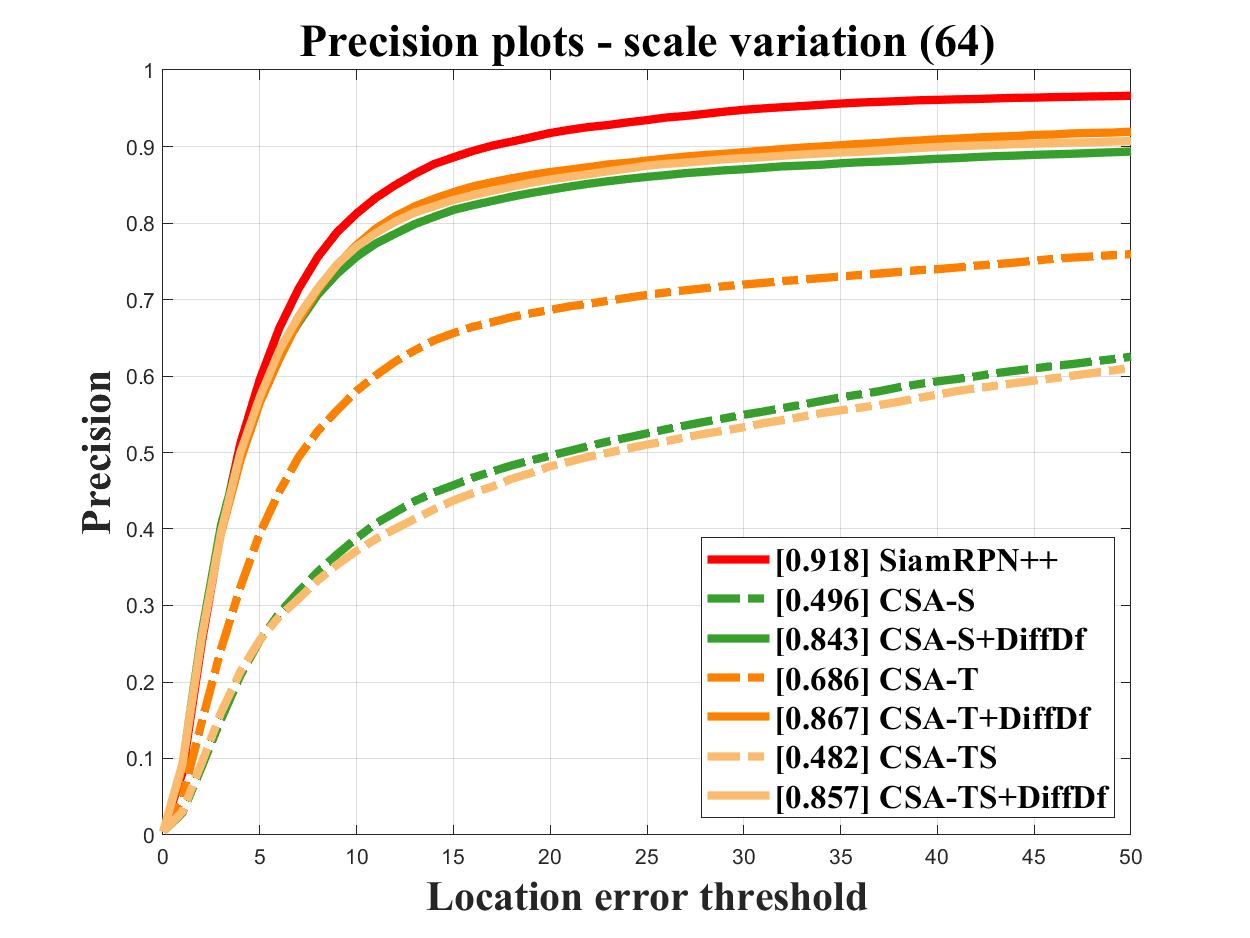}}
	\end{center}
	\caption{Precision plots of our proposed DiffDf method on defending the SiamRPN++ tracker\cite{r2} against the white-box CSA attacks\cite{r22} on the OTB2015 dataset\cite{r39}.}
	\label{fig:5}
\end{figure}

The results on the OTB2015\cite{r39} dataset also validate the effectiveness of our proposed defense method. After the CSA-S attack, the success rate of SiamRPN++ dropped to 0.346, but after applying our defense method, the success score rebounded to 0.658, with a recovery rate of 94.4\%. In particular, for the CSA-T attack, the precision score after defense reached 0.871, 96.3\% of the original baseline of 0.917, confirming the significant effect of the proposed method in suppressing adversarial perturbations in the feature space.

Additionally, to further analyze the performance of the proposed defense method under different challenge attributes in this dataset, Figs.\ref{fig:4} and \ref{fig:5} show the success and precision plots for each attribute. CSA attacks significantly weakened the tracking performance of the original SiamRPN++, especially under the CSA-TS attack, where the success score drastically dropped, and stable tracking ability was almost lost. After applying DiffDf, the tracker achieved significant performance recovery in most scenarios, with an average success score improvement of 20\% to 30\%, which fully demonstrates its good defensive adaptability. Specifically, in scenarios where the target appearance and shape change, such as fast motion, motion blur, and deformation, DiffDf effectively suppresses the incorrect feature matching caused by perturbations through its structural reconstruction mechanism. In scenarios like occlusion and out-of-view, DiffDf leverages its semantic consistency preservation strategy to retain historical semantic information of the target object, thus mitigating the risk of tracking interruption and achieving stable recovery. In scenarios where scale variation and low resolution degrade image quality, DiffDf maintains high success scores and tracking accuracy, significantly improving precision over the attacked baseline. Moreover, DiffDf consistently enhanced performance across all three attacks (CSA-S, CSA-T, and CSA-TS). Whether the disturbance occurred in the target template, search region, or both, the defense effect remained stable and reliable. This further demonstrates that the proposed DiffDf method has excellent generalization ability to attack locations, enabling robust adaptation to diverse perturbation scenarios and maintaining consistent defense effectiveness.

In the long-term tracking scenario of LaSOT\cite{r40}, the success score of the original SiamRPN++ was 0.496. After a white-box CSA attack on both the target template and the search region (CSA-TS), the success score dropped significantly to 0.168, decreasing 66.1\%. After applying the DiffDf defense method, the success score recovered to 0.466, reaching 93.9\% of the baseline. It is worth emphasizing that the defense effect was even more significant in the case of single-attack scenarios (\ie CSA-S and CSA-T). The success scores for CSA-S+DiffDf and CSA-T+DiffDf were 0.484 and 0.492, respectively, only about 2.4\% and 0.8\% lower than the baseline. Meanwhile, the P$_{\text{norm}}$ score reached 0.557 and 0.565, respectively, close to the baseline of 0.569.

\subsection{Visualized results}

\begin{figure}[h!]
	\begin{center}
		\includegraphics[width=\linewidth]{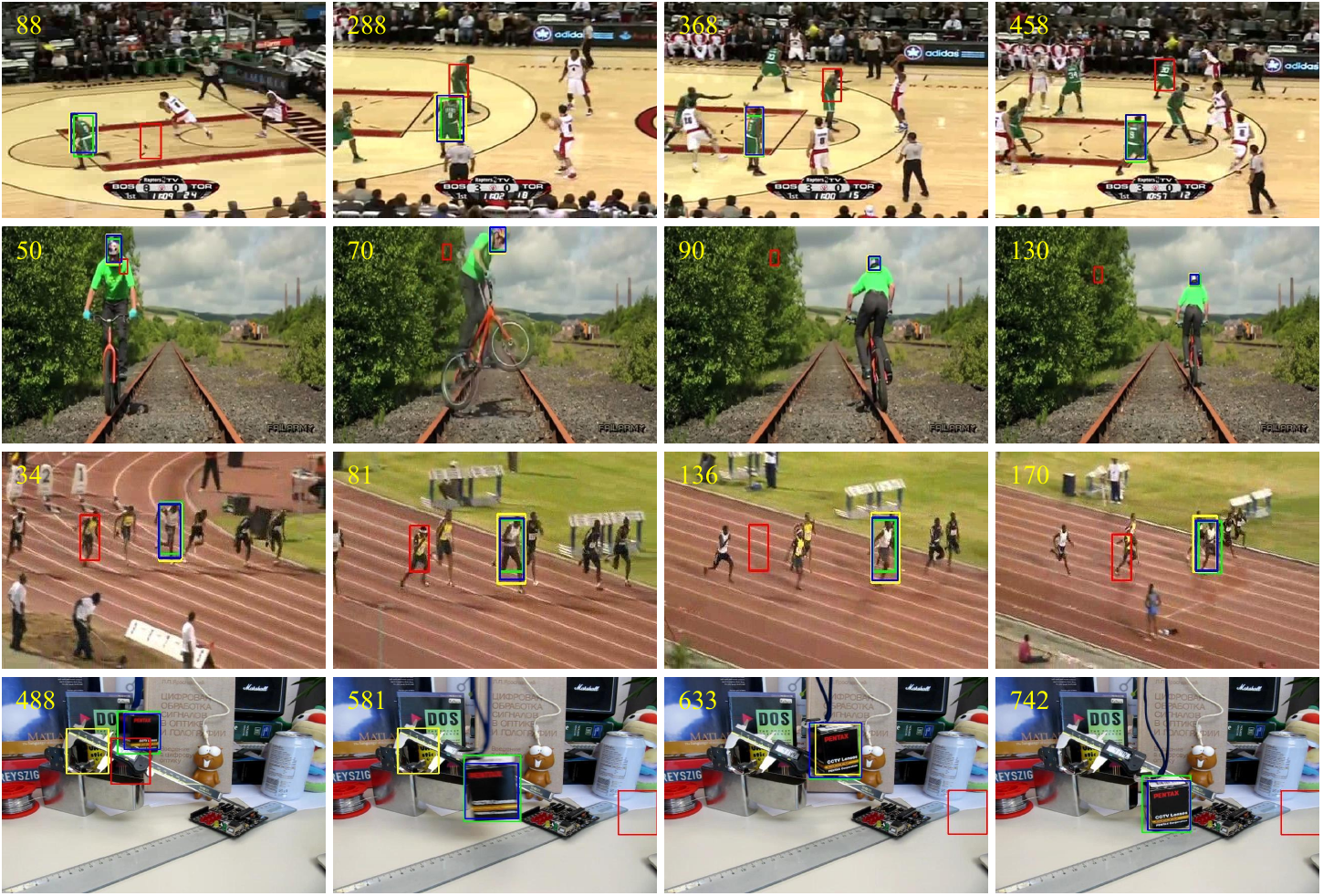}
	\end{center}
	\caption{Visualization of tracking results of SiamRPN++\cite{r2} on original video frames (\textcolor{yellow}{\textbf{yellow}} bounding boxes), after the white-box CSA-S attack (\textcolor{red}{\textbf{red}} bounding boxes), and after the DiffDf defense (\textcolor{blue}{\textbf{blue}} bounding boxes) from the OTB2015 dataset\cite{r39}. Ground-truth bounding boxes are also provided in \textcolor{green}{\textbf{green}}.}
	\label{fig:6}
\end{figure}

To more intuitively demonstrate the defense effect of the proposed DiffDf method in complex tracking scenarios, we selected several representative video frames for visualization. Fig.\ref{fig:6} shows the tracking results of SiamRPN++\cite{r2} in the OTB2015 dataset\cite{r39} without any attack (yellow), after being attacked by the white-box CSA-S method (red), and after defense with the DiffDf method (blue), compared with the ground truth (green). After the attack, SiamRPN++ exhibited noticeable tracking drift or failure. However, after defense with the diffusion model, the tracking results defended by the DiffDf method are closer to the ground truth in most frames, indicating that the method effectively mitigates the disruption of adversarial perturbations on target-specific feature representation, and the tracking results are close to or restored to the level under no attack.

It is worth mentioning that in the last row of Fig.\ref{fig:6}, the original tracking results still showed slight drift even when not attacked, while the tracking results defended by the DiffDf method aligned even more closely with the ground truth. This illustrates that the progressive denoising process of the diffusion model, while eliminating adversarial perturbations, can further correct small biases left in the tracker, thus achieving better accuracy in some scenarios than in the no-attack case. These visual comparisons show that the original tracker often experiences significant drift due to the attack in complex scenarios such as illumination variation and occlusion. In contrast, the defense results not only suppress the destructive effects of the attack but sometimes even help compensate for the design flaws of the tracker itself, reflecting the robustness of the DiffDf method.

\subsection{Transferability across different tracking architectures}

\begin{figure}[t!]
	\begin{center}
		\subfigure{\includegraphics[width=.49\linewidth]{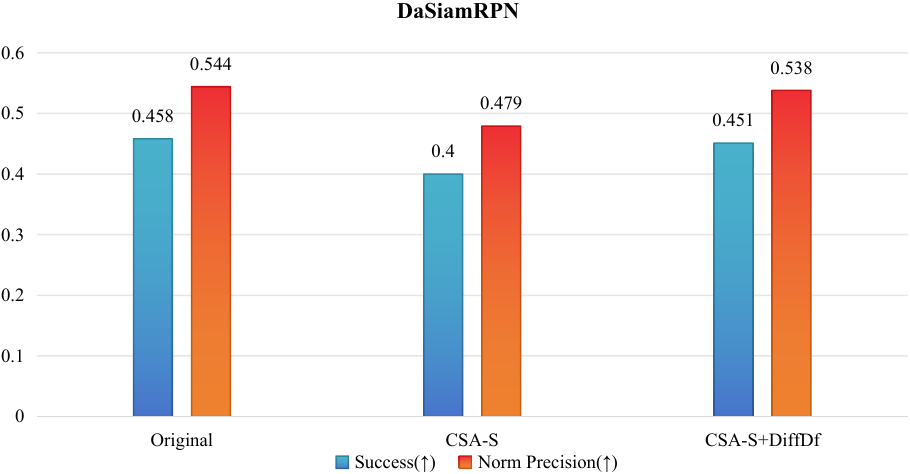}}
		\hspace{0.5em}
		\subfigure{\includegraphics[width=.49\linewidth]{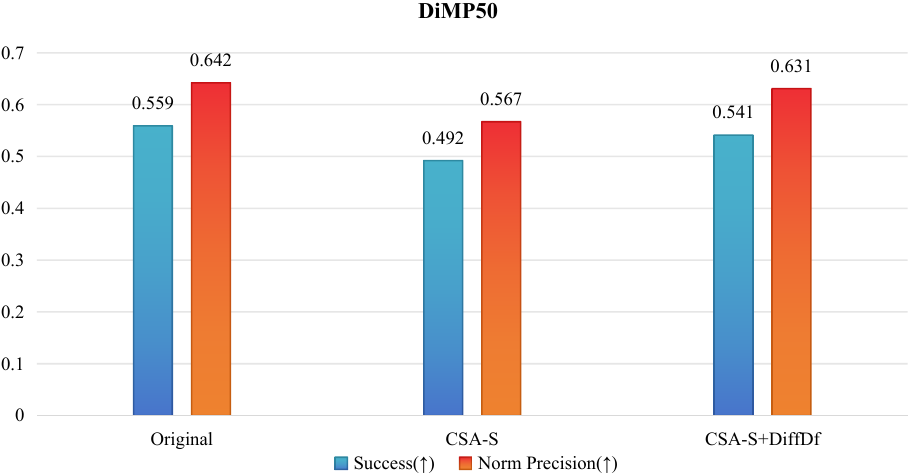}}
	\end{center}
	\caption{The transferability validation results of our proposed DiffDf defense method on trackers with different architectures (DaSiamRPN\cite{r36} and DiMP50\cite{r3}) on the LaSOT\cite{r40} dataset.}
	\label{fig:7}
\end{figure}

To comprehensively evaluate the effectiveness of the proposed defense method and its transferability across different tracking architectures, we conducted experiments on representative trackers DaSiamRPN\cite{r36} and DiMP50\cite{r3} using the LaSOT\cite{r40} dataset, with results shown in Fig.\ref{fig:7}.

For DaSiamRPN, the white-box CSA-S attack\cite{r22} significantly degraded its tracking performance, with the success and normalized precision dropping to 0.400 and 0.479, respectively. After applying the proposed defense method (CSA-S+DiffDf), performance improved significantly, with success and normalized precision recovering to 0.451 and 0.538, respectively. This indicates that the proposed DiffDf method can effectively mitigate the impact of the CSA attack and enhance the robustness of DaSiamRPN. A similar trend was observed with the DiMP50 tracker. The CSA-S attack\cite{r22} caused the success and normalized precision of DiMP50 to decrease to 0.492 and 0.567, respectively. However, after defending with the proposed DiffDf method, these metrics improved to 0.541 and 0.631, respectively. This further validates the transferability and effectiveness of the proposed method across different tracking architectures.

\subsection{Compared with existing defense methods}

\begin{table}[h]
	\centering
	\caption{Performance comparison of different defense methods using the SiamRPN++\cite{r2} tracker under the white-box CSA-T attack\cite{r22} on the VOT2018\cite{r38} dataset.}
	\label{tab:3}
		\begin{tabular}{l|cccc}
			\toprule
			Methods       & Accuracy $\uparrow$ & Robustness $\downarrow$ & Lost number $\downarrow$ & EAO $\uparrow$   \\
			\midrule
			\rowcolor{lightgray!40}
			Original   & 0.609    & 0.276      & 59          & 0.374 \\
			CSA-T         & 0.541    & 1.147      & 245         & 0.123 \\
			\rowcolor{green!20}
			CSA-T+Gaussian  & 0.568    & 0.684      & 146         & 0.195 \\
			\rowcolor{green!20}
			CSA-T+Median & 0.578    & 0.730      & 156         & 0.182 \\
			\rowcolor{yellow!40}
			CSA-T+DiffDf & 0.596    & 0.300      & 64          & 0.341 \\
			\bottomrule
		\end{tabular}%
\end{table} 

As shown in Table \ref{tab:3}, we conducted a comprehensive experimental comparison of the proposed DiffDf defense method with other common denoising methods to evaluate the defense performance against the white-box CSA-T attack\cite{r22} (attack the target template) using the SiamRPN++\cite{r2} tracker on the VOT2018\cite{r38} dataset. The CSA-T attack significantly degraded the performance of SiamRPN++, with accuracy dropping to 0.508, a reduction of approximately 16\% compared to the baseline of 0.604. Meanwhile, the robustness worsened to 1.456, and the lost number surged to 311, nearly 3.6 times higher than the baseline. The EAO dropped sharply to 0.090, only 24\% of the baseline.

After applying traditional Gaussian and median filtering methods, although some of the negative effects of the attack were mitigated, the recovery remained limited. Specifically, the accuracy of the Gaussian and median filtering methods increased to 0.568 and 0.578, respectively, improving by 12\% and 14\% compared to the attacked results but still 6\% lower than the baseline. The robustness showed limited improvement, with the lost number remaining high at 146 and 156, respectively, twice the baseline. The EAO metric only recovered to 0.195 and 0.182, still more than 48\% lower than the baseline. In contrast, the proposed DiffDf method demonstrated significantly better defense capabilities than traditional denoising methods. Specifically, the accuracy of the defense improved to 0.596, only about 1.3\% lower than the baseline. The robustness significantly improved to 0.300, recovering 96\% of the baseline, with the lost number decreasing to 64, only 20\% of the attack result. Moreover, the EAO increased to 0.341, recovering 91\% of the baseline, significantly outperforming other denoising methods. This demonstrates that the DDPM-based defense method proposed in this study can effectively resist the impact of adversarial attacks, exhibiting superior effectiveness.

\begin{figure}[t!]
	\begin{center}
		\subfigure{\includegraphics[width=0.9\linewidth]{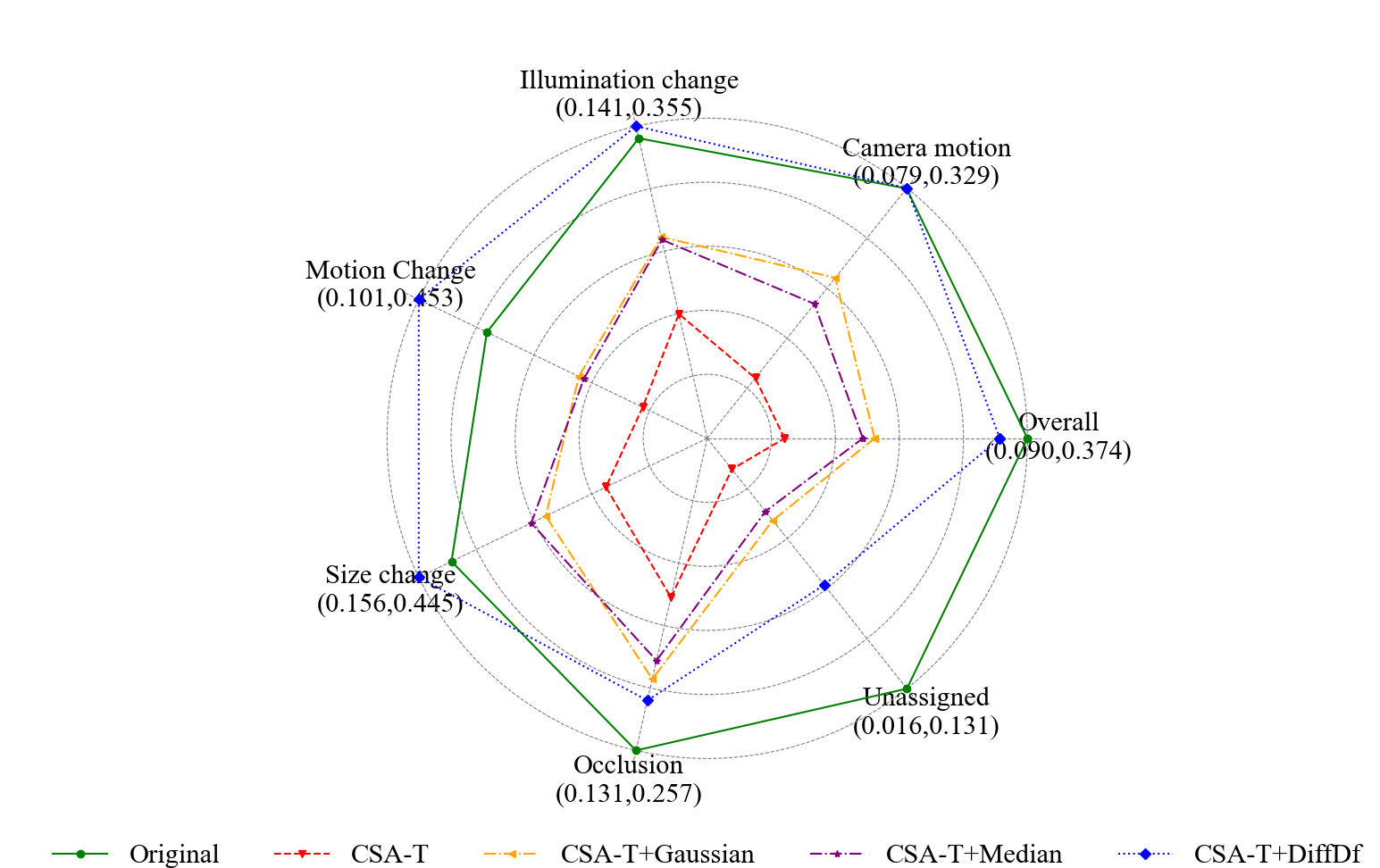}}
	\end{center}
	\caption{Performance comparison of different defense methods using SiamRPN++\cite{r2} tracker in different tracking scenarios (camera motion, illumination change, motion change, size change, occlusion, and unassigned) of the VOT2018\cite{r38} dataset.}
	\label{fig:8}
\end{figure}

Fig.\ref{fig:8} shows the performance under various challenging tracking scenarios on the VOT2018 dataset. The DiffDf method exhibits clear advantages under all scenarios, especially in motion change and size change scenarios, where its performance metrics reach 0.453 and 0.445, outperforming the baseline by 31\% and 13\%, respectively. This indicates that the DiffDf method can effectively counteract adversarial perturbations interference and further enhance tracking performance under specific challenging scenarios. In the camera motion and illumination change scenarios, the DiffDf method's performance is 0.329 and 0.355, respectively, almost on par with the baseline (0.328 and 0.341), demonstrating stable defense effectiveness. In the most challenging occlusion scenario, although the DiffDf method's performance is slightly lower than the baseline, it still significantly outperforms traditional Gaussian and median filtering methods and the CSA-T attack. Specifically, the EAO in the occlusion scenario for the DiffDf method is 0.216, which improves by 65\% compared to the CSA-T attack and is higher than Gaussian filtering of 0.198 and median filtering of 0.183. In the unassigned scenario, the EAO of the DiffDf method reaches 0.077, significantly outperforming other methods.

\begin{figure}[t!]
	\begin{center}
		\subfigure{\includegraphics[width=\linewidth]{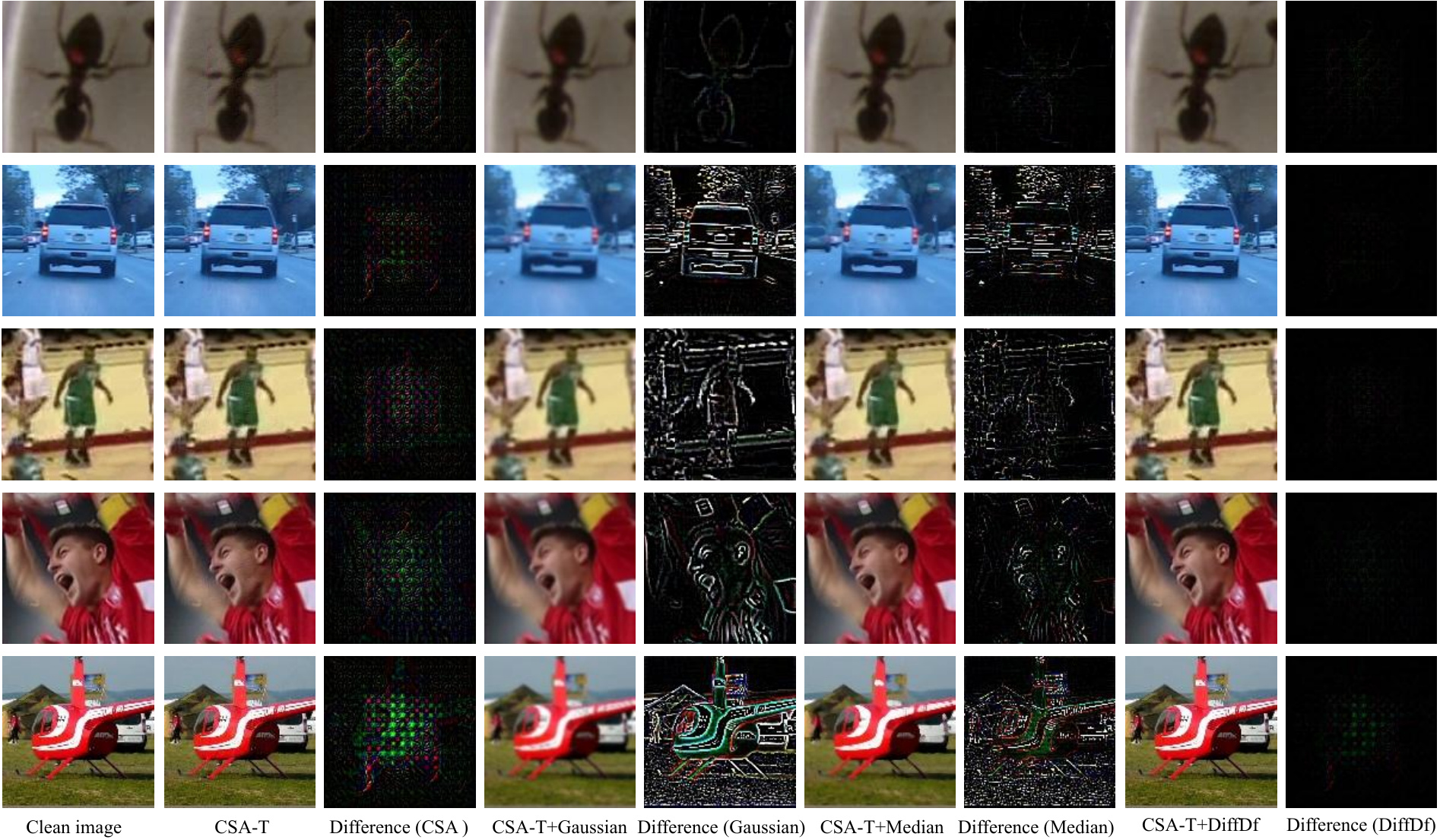}}
	\end{center}
	\caption{Visual comparison of the effects of different defense methods on the target templates of the SiamRPN++\cite{r2} tracker.}
	\label{fig:9}
\end{figure}

We also visualized the target templates before and after the CSA-T attack, as shown in Fig.\ref{fig:9}. The first column shows the original clean image, while the second and third columns show the images after the CSA-T attack and their difference maps (magnified 10 times for better visualization), respectively. The CSA-T attack introduces noticeable noise and structural perturbations into the image. While Gaussian and median filtering methods somewhat reduce the noise intensity, the restored images still exhibit noticeable structural differences, indicating that traditional methods, while suppressing local noise, struggle to preserve the global structural information of the image. In contrast, the proposed DiffDf method (the last two columns) demonstrates superior visual quality in image recovery compared to traditional methods. The difference map shows that the DiffDf method nearly eliminates structural disturbances, with pixel differences significantly reduced. This is mainly due to the multiple constraints applied during the denoising process, effectively balancing pixel-level similarity with structural fidelity, thereby preserving the original structural information of the image while removing adversarial perturbations.

\subsection{Efficiency analysis}

To further explore the feasibility of deploying the DiffDf method in practice, we evaluated the number of parameter, computational overhead (GFLOPs), and inference speed (FPS). The experiment compared the performance of the SiamRPN++\cite{r2} tracker, the tracker after being attacked by the CSA-S attack\cite{r22} (attack the search region), and the tracker defended using the DiffDf method. The detailed experimental results are shown in Table \ref{tab:4}.

\begin{table}[h]
	\centering
	\caption{Comparison of the number of parameters, computational overhead, and inference speed of the original SiamRPN++\cite{r2} tracker, after being attacked by the CSA-S method\cite{r22}, and after being defended using the DiffDf method.}
	\label{tab:4}
		\begin{tabular}{l|rrr}
			\toprule
			Methods    & \#Parameter (M) & GFLOPs & FPS \\
			\midrule
			\rowcolor{lightgray!40}
			Original   & 32.70            & 59.60                           & 60.59                 \\
			CSA-S        & 87.11           & 92.29                           & 39.13                 \\
			\rowcolor{yellow!40}
			CSA-S+DiffDf & 200.78          & 340.47                          & 34.01          \\
			\bottomrule
		\end{tabular}%
\end{table}

The original SiamRPN++ tracker runs at 60.59 FPS. After the CSA attack, the number of parameters significantly increased to 87.11 M, computational overhead rose to 92.29 GFLOPs, and the inference speed dropped noticeably to 39.13 FPS. This is primarily because the CSA-S attack directly interferes with the similarity matching process in the search region, increasing the computational complexity and causing a significant reduction in inference speed. After applying the proposed DiffDf method for defense, the parameter count increased to 200.78 M, the computational overhead rose to 340.47 GFLOPs, and the inference speed slightly dropped to 34.01 FPS. This is mainly because the diffusion model requires multiple iterative computations, and each time step significantly adds to the computational complexity. However, the DiffDf method itself, as a plug-and-play preprocessing module, does not require any modification to the existing tracker architecture; its defense process is entirely performed at the image input stage. Nevertheless, our DiffDf method can still achieve real-time inference speeds above 30 FPS on mid-range GPUs, demonstrating high real-time performance and feasibility in practical deployment.

\subsection{Ablation studies on the loss functions}

\begin{table}[h]
	\centering
	\caption{Results of ablation studies on different combinations of loss functions using the SiamRPN++\cite{r2} tracker under the white-box CSA-S attack\cite{r22} on the VOT2018\cite{r38} dataset.}
	\label{tab:5}
		\begin{tabular}{c|ccccc}
			\toprule
			\multirow{2}{*}{Tracker}   & \multicolumn{3}{c}{Loss functions} &  & VOT2018 \\ \cline{2-6}
			& $\mathcal{L}_\text{pixel}$         & $\mathcal{L}_\text{semantic}$         & $\mathcal{L}_\text{ssim}$        &  & EAO $\uparrow$    \\
			\midrule
			\multirow{4}{*}{SiamRPN++\cite{r2}} & $\surd$           & $\times$         & $\times$         &  & 0.302   \\
			& $\surd$          & $\surd$         & $\times$         &  & 0.318   \\
			& $\surd$          & $\times$       & $\surd$           &  & 0.320   \\
			& $\surd$          & $\surd$         & $\surd$           &  & 0.332   \\
			\bottomrule
		\end{tabular}%
\end{table}

To further analyze the impact of the loss strategies used in the DiffDf method on defense performance, we conducted a detailed ablation experiment on the SiamRPN++\cite{r2} tracker under the white-box CSA-S attack\cite{r22} using the VOT2018\cite{r38} dataset. The results are shown in Table \ref{tab:5}.

When only pixel-level reconstruction loss was used, the EAO was 0.302, indicating that while pixel-level reconstruction can restore the basic information of the image, it lacks effective constraints on structural consistency and high-level semantic information, resulting in limited defense effectiveness. After adding the structural similarity loss, the EAO improved to 0.320, suggesting that structural similarity is vital in restoring image quality. Additionally, after incorporating the semantic consistency loss, the EAO further increased to 0.318, reflecting the effectiveness of semantic-level feature constraints in defending against adversarial attacks. When combining pixel-level reconstruction loss, structural similarity loss, and semantic consistency loss, DiffDf achieved the best defense performance with an EAO of 0.332, indicating that the joint effect of these three losses, under multi-scale and multi-level constraints, can significantly improve both defense performance and image quality.

\subsection{Discussions}

Although the proposed DiffDf method demonstrates strong performance in defending against various types of adversarial perturbations and exhibits appealing structural generality and practical deployability, it still faces the following limitations that need to be addressed in future research.

DiffDf relies on a multi-step reverse diffusion process to progressively remove perturbations, constructing a denoising learning path. Compared to traditional image filtering or shallow CNNs, the inference process involves a large number of intermediate representations and feature calculations. While we employed U-Net architecture compression and a scheduling mechanism to enhance efficiency, and our practical implementation maintains a frame rate of over 30 FPS---satisfying real-time requirements for visual tracking---deployment in resource-constrained environments still presents challenges. Future work can explore lightweight diffusion mechanisms combined with DDIM, FastSampler, or downsampling strategies to reduce computational costs while maintaining robustness.

DiffDf serves as a plug-and-play image-level preprocessing module, its primary advantage lies in its zero-intrusiveness to the tracker architecture and strong adaptability, allowing it to be directly integrated into various mainstream black-box trackers. However, this structural decoupling means that DiffDf cannot be aware of specific downstream optimization goals during training, such as classification confidence or bounding box regression biases. This may result in denoised images that are perceptually clearer but may not achieve optimal performance for the tracking task. Future work could explore end-to-end defense methods for visual tracking based on feature sharing or joint training, improving the consistency between feature semantics and task objectives.

Current training and testing of DiffDf are based on standard input sizes and have not been thoroughly validated in lightweight diffusion architectures or non-natural image domains (such as infrared, remote sensing, etc.). The transferability and adaptation of DiffDf in complex noise backgrounds, varying distributions, and cross-modal data still need further exploration. The diffusion model is also sensitive to the number of timesteps, scheduling strategies, and normalization methods, which may lead to nonlinear performance degradation depending on the input domain. Therefore, enhancing the generalization ability across domains, resolutions, and multi-task environments is an important direction for future work.

\section{Conclusion}\label{sec:5}

This paper presents an adversarial defense method based on the DDPM named DiffDf. Specifically, the DiffDf method significantly enhances the robustness of existing visual tracking methods in adversarial attack scenarios through a multi-scale optimization strategy involving pixel-level reconstruction, semantic consistency constraints, and structural similarity losses. Extensive experiments on several mainstream benchmark datasets validate the excellent defense capabilities of DiffDf against various types of attacks (\eg attack target template or search region). Moreover, DiffDf demonstrates amazing transferability, showing stable and outstanding performance across different tracking architectures (such as SiamRPN++, DaSiamRPN, and DiMP50), significantly outperforming traditional filtering denoising methods. Although DiffDf demonstrates significant defense effectiveness, several challenges remain that are worth addressing. First, the computational complexity of the diffusion model is relatively high, which may limit its deployment in resource-constrained devices. Second, as an independent preprocessing module, DiffDf has not yet been jointly optimized with the target tracker, thus failing to fully exploit the potential for collaborative training and learning between defense and tracking tasks. These issues provide directions for future research, such as developing efficient network architectures, achieving defense-tracking collaborative optimization, and enhancing the general deployment capability.

\bibliographystyle{num}
\bibliography{bibliography}


\end{document}